\documentclass[10pt,twocolumn,letterpaper]{article}

\PassOptionsToPackage{table}{xcolor}
\usepackage[pagenumbers]{cvpr}
\usepackage{multirow}

\definecolor{cvprblue}{rgb}{0.21,0.49,0.74}
\usepackage[pagebackref,breaklinks,colorlinks,allcolors=cvprblue]{hyperref}

\title{
$\boldsymbol{\widetilde{\varepsilon}}$: Does Initial Noise Induce Directional Bias in\\ Vision-Language-Action Models?
}

\author{%
Junmyeong Lee$^{1}$,
Dongmin Shin$^{1,2}$,
Wooseok Jeon$^{1}$,
JunGyu Lee$^{1}$,
Seunghyun Shin$^{3}$,
Hae-Gon Jeon$^{1,}$%
\setcounter{footnote}{1}%
\thanks{\,Corresponding author.\newline
\makebox[1.8em][r]{\textsuperscript{1}}\,Yonsei University\newline
\makebox[1.8em][r]{\textsuperscript{2}}\,Korea Electronics Technology Institute (KETI)\newline
\makebox[1.8em][r]{\textsuperscript{3}}\,Gwangju Institute of Science and Technology (GIST)}%
\\[0.5em]
\url{https://june65.github.io/Epsilon-Tilde}%
}

\begin{document}
\maketitle
\begin{abstract}
Vision-language-action models (VLAs) have widely adopted diffusion-based action generation for continuous robot motions, enabling fine-grained control across diverse tasks.
However, recent studies have shown that their performance is highly sensitive to initial noise, with resulting action errors that make recovery difficult during execution.
Existing approaches search for favorable noise initializations to improve performance, yet the action expert itself remains sensitive to noise.
In this work, we reveal that initial noise induces a consistent directional bias in action space that closely aligns with the noise’s temporal mean.
We thus introduce a simple yet effective framework that improves robustness to initial noise through interventions during both training and inference.
During training, we enforce consistent action predictions from antithetic noise pairs \((\epsilon, -\epsilon)\) and, at inference, remove the noise’s temporal mean.
Our method substantially reduces action deviation and improves task performance across simulation benchmarks.
In real-robot experiments, our method increases the average task success rate of \(\pi_{0.5}\) from 40\% to 68\%, with more stable and consistent execution.
\end{abstract}

\begin{figure}[t]
  \centering
   \includegraphics[width=1\linewidth]{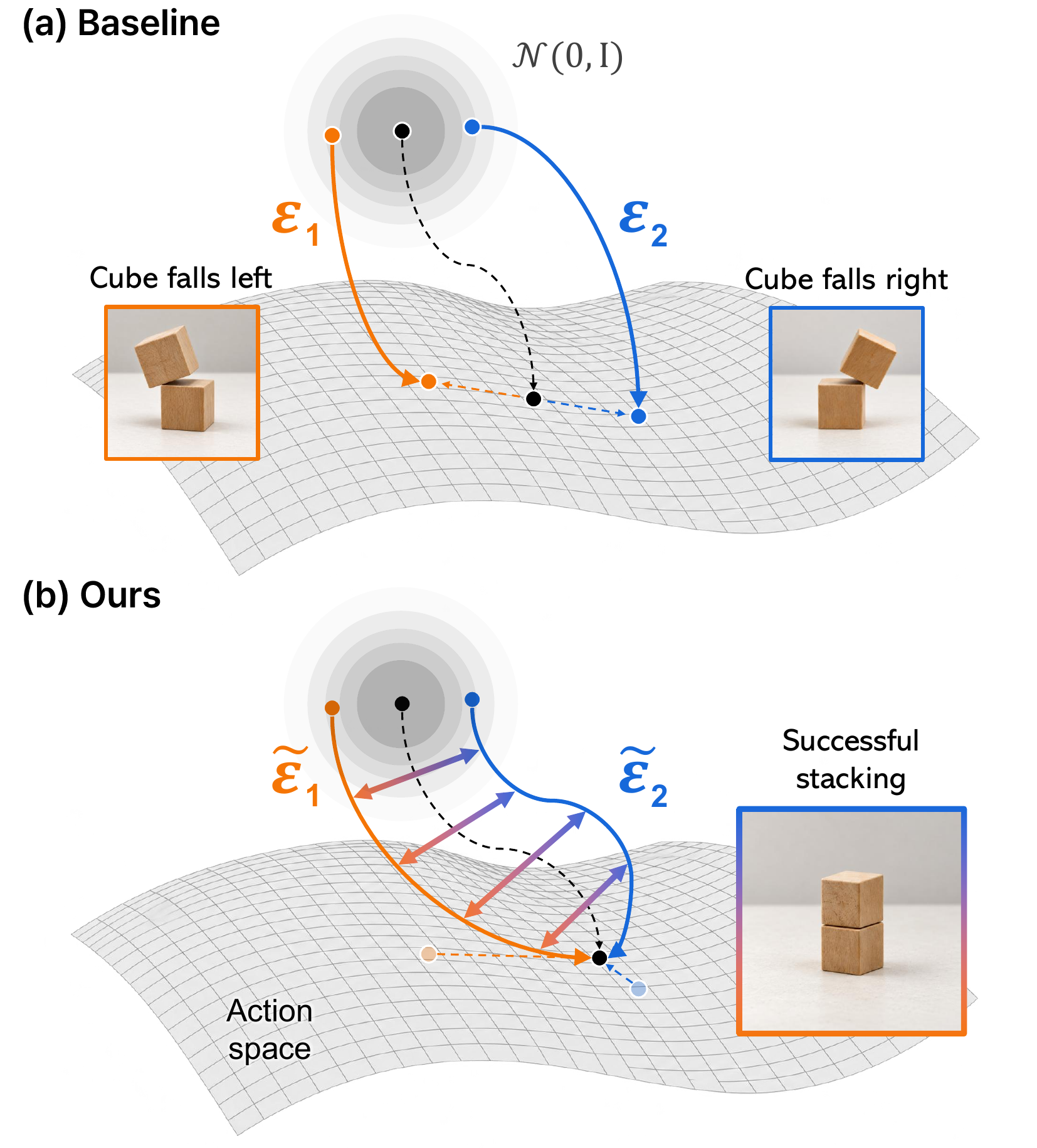}
    \caption{\textbf{Initial noise sensitivity and directional bias.} (a) Different initial noise samples $\epsilon_1$ and $\epsilon_2$ can induce directional action deviations that cause stacking failures. (b) Our method reduces these deviations, promoting more consistent execution across noise samples and enabling stable stacking.}
   \label{fig:fig_1}
\end{figure}

\section{Introduction}
\label{sec:intro}

Recent advances in diffusion models~\cite{ho2020denoising,song2020denoising,rombach2022high,lipman2022flow} have enabled vision-language-action models (VLAs) to generate continuous action sequences for smooth robot control~\cite{chi2025diffusion,black2024pi,pmlr-v305-black25a,bjorck2025gr00t}.
As a result, VLAs can perform an increasingly broad range of tasks, from simple pick-and-place operations to long-horizon manipulation~\cite{kim2024openvla,kim2025fine,cen2025worldvla,bu2025univla,zheng2026x,shukor2025smolvla,wen2025dexvla,team2025gemini}.
However, recent studies identify initial noise as a key factor contributing to inconsistent task performance~\cite{patil2026you,yang2025steering}.
Since a single inaccurate motion can significantly alter the next observation and trigger cascading failures, robustness to initial noise is crucial for consistent manipulation. Also, recovery can be particularly difficult when such errors change the physical state of objects, as illustrated in~\cref{fig:fig_1}~(a), where a cube falls during stacking. Therefore, improving robustness to initial noise is essential for reliable real-world deployment of VLAs.

Prior works have explored two main approaches to address initial noise sensitivity in diffusion-based policies.
One research direction is to improve manipulation performance by searching for favorable initial noise or evaluating multiple generated action candidates~\cite{patil2026you,wagenmaker2025steering,kwok2025robomonkey,yang2025steering}.
Another line of work learns the noise distribution or modifies the noise-to-action mapping during training~\cite{dai2026should,zhong2026noise,liu2025sample}.
However, both approaches often require additional computation during inference or training for task-specific adaptation.
Furthermore, they do not fully explain the systematic changes that initial noise induces in action space, and the resulting policies remain sensitive to noise initialization.

To address this limitation, we analyze how initial noise affects action generation.
We find that it inherently induces directional deviations in generated actions, which we refer to as \textit{directional bias}.
We empirically reveal that the same noise sample produces similarly directed deviations despite different observations and language instructions, while antithetic noise pairs tend to induce biases in opposite directions.
These observations suggest that the initial noise is a primary driver of directional bias.
Further analysis reveals that this direction closely aligns with the temporal mean of the initial noise.
By assuming the action distribution as Gaussian, we theoretically show that the directional bias remains aligned with the initial noise’s temporal mean.
These findings motivate a method that reduces directional bias for more stable action generation across noise initializations.

Guided by these insights, we propose intuitive solutions that address directional bias at both training and inference time.
Specifically, during training, we enforce consistency between action predictions obtained from antithetic noise pairs to reduce the effect of initial noise.
This encourages the action expert to generate more consistent actions across different noise initializations.
At inference, we directly subtract the temporal mean from each sample and use it to initialize the action generation (see \cref{fig:fig_1}~(b)).

Through comprehensive evaluations, we confirm that reductions in noise sensitivity are generally accompanied by improved task performance, supporting the importance of addressing the directional bias.
These gains are achieved through simple corrections that can be readily applied to a variety of VLA models.
The improvements also extend to real-robot execution, where our method increases the average task success rate of \(\pi_{0.5}\) from 40\% to 68\%, while improving motion stability.

\section{Related Work}
\label{sec:2_rel_work}

\subsection{Improving Diffusion-Based Robot Policies}
\noindent\textbf{Inference-time intervention.}\quad
During the denoising process, various intervention methods can be applied to improve the stability of generated actions.
These approaches can be broadly categorized into three groups based on where the intervention occurs: initial noise selection, intermediate denoising guidance, and final action selection. \textit{(a) Initial Noise Selection.} Favorable noise initializations steer a frozen policy toward desirable actions~\cite{simsir2026lagrangian, ho2026start, tang2026improving}. DSRL~\cite{wagenmaker2025steering} uses reinforcement learning to learn a policy that selects initial noise, while Golden Ticket~\cite{patil2026you} searches for a high-performing noise vector reusable across observations. \textit{(b) Intermediate Denoising Guidance.} Guidance redirects intermediate samples toward task objectives~\cite{yang2026guided, sirigiri2026bayesfp, park2026acg, zhou2026test, du2026dynaguide}. VLS~\cite{liu2026vls} uses a vision-language model (VLM) to generate rewards that reflect spatial and semantic requirements, while Hume~\cite{song2025hume} evaluates partially denoised candidates and completes selected proposals through cascaded denoising. \textit{(c) Final Action Selection.} Candidate evaluation selects among sampled actions using value estimates, verification, or distributional criteria~\cite{nakamoto2024steering, kwok2025robomonkey, jang2026verifier, nguyen2026self, zhao2026action, xu2026vgas}. KeyStone~\cite{dai2026geometry} selects the medoid of the dominant cluster, while TACO~\cite{yang2025steering} uses a pseudo-counting module to select actions. However, such interventions often introduce additional preparation or inference costs due to noise search or repeated candidate evaluation.

\noindent\textbf{Training-time source and policy design.}\quad
Another line of work improves diffusion-based robot policies by redesigning their action generation framework. ResVLA~\cite{zhong2026noise} anchors generation to predicted motion intent through a residual diffusion bridge, while WarmPrior~\cite{kang2026warmprior} constructs a temporal source prior from previous action predictions. Relevant methods incorporate sources conditioned on the state~\cite{dai2026should}, learned motion priors~\cite{machado2026flowing}, or stochastic dynamics conditioned on observations~\cite{liu2025sample}. Also, GeCO~\cite{zhang2026generative} uses convergence to guide adaptive refinement, while BehaviorVLA~\cite{hu2026abstraction} improves policy representations by modeling long-horizon behavior. However, these approaches may be sensitive to errors in predicted priors and require changes to existing model architectures or generation processes.
\begin{figure*}[t]
  \centering
  \includegraphics[width=1\linewidth]{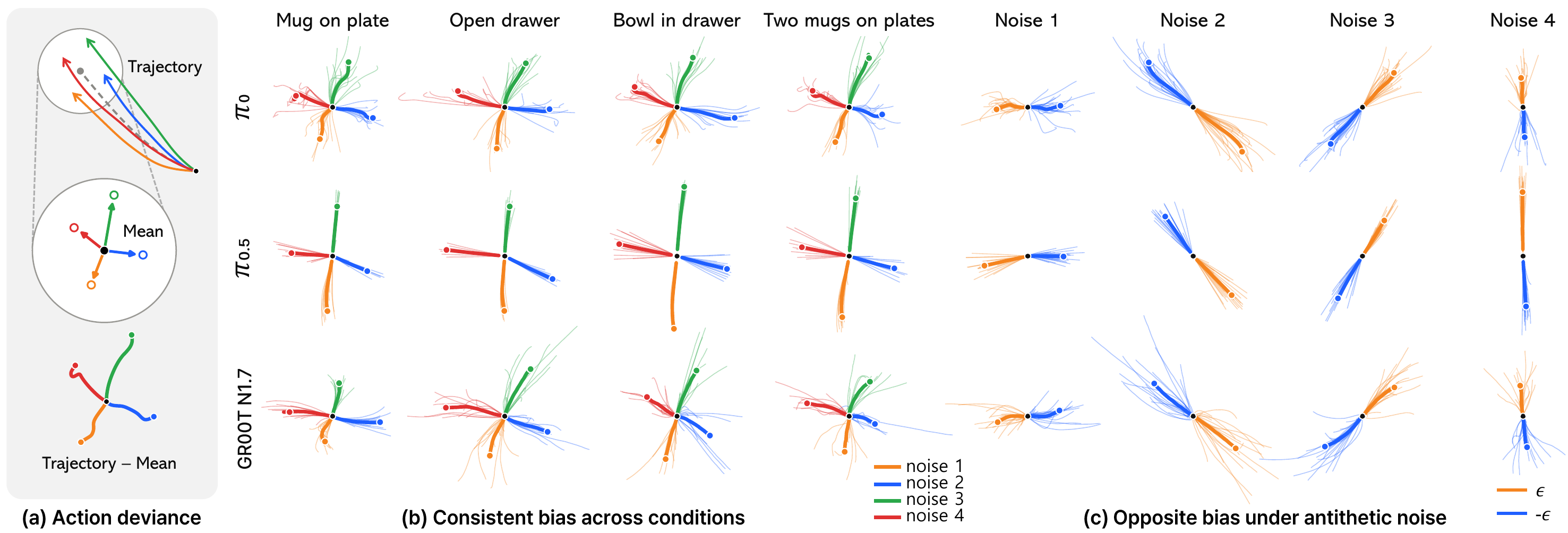}
    \caption{\textbf{Observations.} Across $\pi_0$, $\pi_{0.5}$, and GR00T N1.7, we visualize (a) action deviations from the mean trajectory, showing that (b) the same noise induces similar bias directions across conditions and (c) noise negation tends to reverse them. Colors distinguish noise samples in (b), while orange and blue denote $\epsilon$ and $-\epsilon$ in (c).}
  \label{fig:fig_2}
\end{figure*}

\section{Does Initial Noise Induce Directional Bias?}
\subsection{Preliminaries: Flow-Based Action Generation}
\label{subsec:action_generation_background}
Diffusion-based action experts generate action sequences from initial noise, conditioned on observations and language instructions. Let $c$ denote this conditioning input and $x_0\in\mathbb{R}^{H\times d}$ a target action sequence with horizon $H$ and action dimension $d$. The initial noise $\epsilon\sim\mathcal{N}(0,I)$ has independent standard Gaussian entries and is sampled independently of $(x_0,c)$. We adopt linear flow matching for the analysis and training objective below.

During training, flow matching learns a velocity field along an interpolation path between the target action sequence and noise:
\begin{equation}
\begin{aligned}
\mathcal{L}_{\mathrm{FM}}(\theta)
&=\mathbb{E}_{x_0,c,\epsilon,t}\!\left[
\left\|
v_\theta(x_t,t,c)-(\epsilon-x_0)
\right\|_F^2
\right],\\
&\quad s.t. \quad x_t
=(1-t)x_0+t\epsilon.
\end{aligned}
\label{eq:fm_training}
\end{equation}
Here, $t\in[0,1]$ denotes the diffusion timestep, $v_\theta$ is the velocity field, and $\|\cdot\|_F$ denotes the Frobenius norm.

At inference, generation starts from noise and integrates an ordinary differential equation (ODE) from $t=1$ to 0:
\begin{equation}
\frac{\mathrm{d}y_t}{\mathrm{d}t}
=
v_\theta(y_t,t,c),
\qquad
y_1=\epsilon.
\label{eq:fm_generation}
\end{equation}
The final action sequence is given by $y_0$. We use $x_t$ for the training interpolation and $y_t$ for the integrated sampling trajectory.

\subsection{Observations}
\label{subsec:empirical_observations}
In this section, we investigate whether initial noise causes directional bias in the outputs of diffusion-based action experts. This means that the same initial noise tends to produce similarly directed action deviations across conditions. To better understand this phenomenon, we first analyze the bias across different observations and language instructions, and examine its response to antithetic noise.
We perform our analysis using \(\pi_{0}\)~\cite{black2024pi}, \(\pi_{0.5}\)~\cite{pmlr-v305-black25a}, and GR00T N1.7~\cite{bjorck2025gr00t} across diverse scene conditions in LIBERO~\cite{liu2023libero}.
For each model and condition, we generate action sequences from multiple noise samples and compute each sequence's deviation from their average.
These action deviations are visualized in~\cref{fig:fig_2}~(a) to examine the directional effects of initial noise.

\noindent\textbf{Observation 1: Consistent bias across conditions.}\quad
We first generate actions under different observations and language instructions with fixed initial noise.
As shown in~\cref{fig:fig_2}~(b), we observe that the action deviations induced by a given noise sample tend to point in similar directions despite changes in the conditioning input.
This suggests that flow matching can propagate noise-specific directional bias into action space across different conditioning inputs.

\noindent\textbf{Observation 2: Opposite bias in antithetic noise.}\quad
We compare action predictions before and after negating each noise sample.
Under the same observation and language instruction, we generate action sequences from (\(\epsilon\)) and (\(-\epsilon\)), and compute their action deviations from the same noise-averaged prediction, as shown in~\cref{fig:fig_2}~(c).
Consistent with the findings of recent work on image diffusion~\cite{jia2026antithetic}, the paired action deviations tend to point in opposite directions in action space.

\noindent\textbf{Hypothesis: Temporal mean of initial noise.}\quad
Based on these observations, we hypothesize that the temporal mean of the initial noise contributes to the directional bias.
Note that the temporal mean refers to the average of the initial noise across action steps, computed separately for each action coordinate.
For a fixed noise sample, it remains unchanged across conditions and reverses sign when the noise is negated, consistent with the two observations above.
Although the noise is sampled from a zero-mean distribution, averaging over a finite number of action steps can yield a nonzero temporal mean.

\subsection{Theoretical Analysis}
\label{subsec:temporal_bias}
To examine this hypothesis, we provide a mathematical derivation of how the temporal mean can induce directional bias in generated actions. We first model the action distribution and derive the relationship between initial noise and generated actions through flow matching. We then show that the temporal mean of the initial noise and directional bias are aligned, explaining the patterns in \cref{subsec:empirical_observations}.

\noindent\textbf{Conditional action model.}\quad
\label{par:conditional_action_model}
For our analysis, we approximate the conditional distribution of the target action sequence for a fixed input $c$ by a Gaussian, following prior works~\cite{levine2016end,paraschos2013probabilistic}.
We assume that the action dimensions are independent and share a temporal covariance:
\begin{equation}
p(x_0|c)
\propto
\exp\!\left[
-\frac12
\left\|
\Sigma_c^{-1/2}(x_0-\mu_c)
\right\|_F^2
\right].
\label{eq:action_gaussian}
\end{equation}
Here, $\mu_c=\mathbb{E}[x_0 | c]$ is the conditional mean action sequence, and\,$\Sigma_c\in\mathbb{R}^{H\times H}$\,captures temporal correlations within action dimensions.\,Both\,quantities may depend on $c$.

\noindent\textbf{Closed-form action generation.}\quad
\label{par:closed_form_generation}
Under the Gaussian action model, the optimal velocity field for the flow-matching objective in~\cref{eq:fm_training} can be derived analytically. Integrating this field exactly from $t=1$ to $t=0$ gives:
\begin{equation}
y_0^\star
=
\mu_c+\Sigma_c^{1/2}\epsilon,
\label{eq:closed_form_generation}
\end{equation}
where $\Sigma_c^{1/2}$ is the principal square root of $\Sigma_c$, and $\star$ denotes this ideal generation process. The derivation is provided in the supplementary material.

\noindent\textbf{Temporal mean and directional bias.}\quad
\label{par:temporal_mean_propagation}
Our goal is to establish the relationship between the temporal mean of the initial noise and the resulting directional bias. We define the temporal mean $\bar{\epsilon}$ of the noise as follows:
\begin{equation}
\bar{\epsilon}
=
\frac{1}{H}\epsilon^\top\mathbf{1}.
\label{eq:noise_temporal_mean}
\end{equation}
Here, $\mathbf{1}\in\mathbb{R}^H$ denotes the all-ones vector.

Similarly, we define the directional bias $b_c(\epsilon)$ as the deviation of $y_0^\star$ from its conditional mean $\mu_c$, averaged across action steps:
\begin{equation}
\begin{aligned}
b_c(\epsilon)
&:=
\frac{1}{H}(y_0^\star-\mu_c)^\top\mathbf{1}\\
&=
\frac{1}{H}\epsilon^\top\Sigma_c^{1/2}\mathbf{1}.
\end{aligned}
\label{eq:generated_temporal_bias}
\end{equation}
The second equality follows from~\cref{eq:closed_form_generation}.
To isolate the contribution of the temporal mean, we average over noise samples that share the same temporal mean:
\begin{equation}
\mathbb{E}[b_c(\epsilon)|\bar{\epsilon},c]
=
\frac{1}{H}
\mathbb{E}[\epsilon|\bar{\epsilon},c]^\top
\Sigma_c^{1/2}\mathbf{1}.
\label{eq:expected_action_bias}
\end{equation}
Since the initial noise has i.i.d. standard Gaussian entries, the noise values at all action steps have the same conditional expectation, equal to the specified temporal mean:
\begin{equation}
\mathbb{E}[\epsilon|\bar{\epsilon},c]
=
\mathbf{1}\bar{\epsilon}^\top.
\label{eq:conditional_noise_mean}
\end{equation}
Substituting this into~\cref{eq:expected_action_bias} gives
\begin{equation}
\begin{aligned}
\mathbb{E}[b_c(\epsilon)|\bar{\epsilon},c]
&=
\underbrace{
\frac{1}{H}\mathbf{1}^\top
\Sigma_c^{1/2}\mathbf{1}
}_{>0}
\bar{\epsilon}.
\end{aligned}
\label{eq:conditional_mean_bias}
\end{equation}
This shows that the conditional expectation of the directional bias is proportional to the temporal mean of the initial noise. The scale depends on the temporal covariance $\Sigma_c$, which may vary across conditions.

\begin{figure}[t]
  \centering
   \includegraphics[width=1\linewidth]{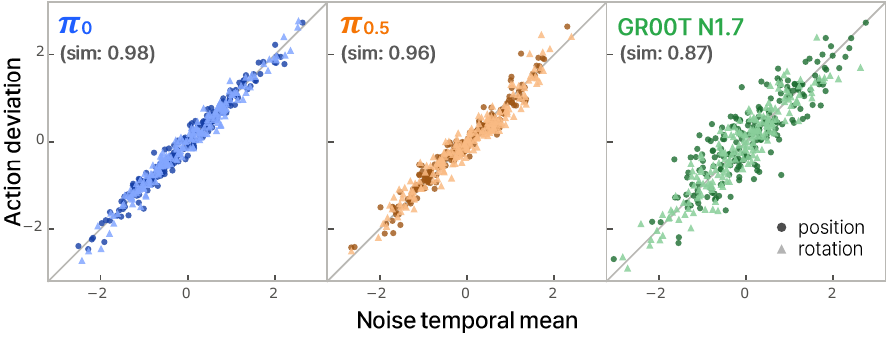}
   \vspace{-5mm}
    \caption{\textbf{Temporal means align with action deviations.} Deviations are averaged across conditions, and the plotted values are standardized within each channel. Circles and triangles denote position and rotation channels.}
   \label{fig:fig_3}
\end{figure}

\begin{figure}[t]
  \centering
   \includegraphics[width=1\linewidth]{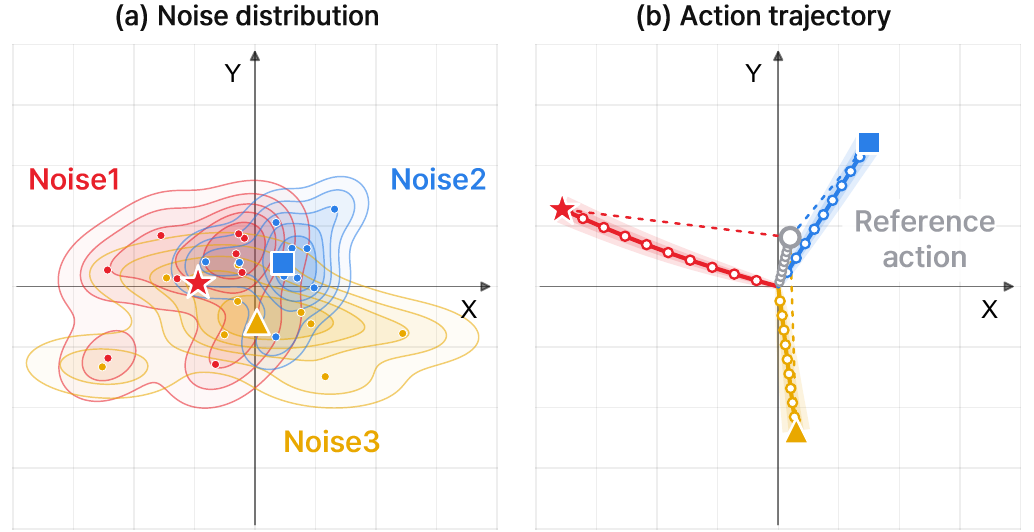}
    \caption{\textbf{Noise distributions and action trajectories.} Noise samples with different temporal means and their corresponding action trajectories.}
   \label{fig:fig_4}
\vspace{-3mm}
\end{figure}

\subsection{Empirical Validation}
We next examine whether the predicted alignment between the initial noise’s temporal mean and directional bias holds across \(\pi_{0}\), \(\pi_{0.5}\), and GR00T N1.7.
Using 64 noise samples and 120 conditioning inputs, we measure the cosine similarity \(\cos\!\left(\bar\epsilon,\mathbb{E}_{c}[b_c(\epsilon)]\right)\) between the temporal mean of each noise sample and its directional bias averaged over conditioning inputs.
As shown in~\cref{fig:fig_3}, the cosine similarity averaged over noise samples is 0.98 for $\pi_0$, 0.96 for $\pi_{0.5}$, and 0.87 for GR00T N1.7.
We further visualize the sampled noise distributions and corresponding action trajectories in~\cref{fig:fig_4}\, where the temporal means of the noise samples align with the resulting directional bias.

\begin{table*}[t]
\centering
\setlength{\tabcolsep}{4pt}
\renewcommand{\arraystretch}{1.15}
\resizebox{\textwidth}{!}{%
\begin{tabular}{
l|
>{\columncolor{green!15}}c
>{\columncolor{green!15}}c|
cccc|cccc|
>{\columncolor{green!15}}c
>{\columncolor{green!15}}c|
ccccc
}
\toprule
\multirow{2}{*}{\vspace{-2mm}Method}
& \multicolumn{1}{c}{
    \multirow{2}{*}{\vspace{-2mm}\shortstack{$M \downarrow$}}}
& \multicolumn{1}{c|}{
    \multirow{2}{*}{\vspace{-2mm}\shortstack{$C \downarrow$}}}
& \multicolumn{4}{c|}{Google Robot (VM)}
& \multicolumn{4}{c|}{Google Robot (VA)}
& \multicolumn{1}{c}{
    \multirow{2}{*}{\vspace{-2mm}\shortstack{$M \downarrow$}}}
& \multicolumn{1}{c|}{
    \multirow{2}{*}{\vspace{-2mm}\shortstack{$C \downarrow$}}}
& \multicolumn{5}{c}{WidowX} \\
\cmidrule(lr){4-7}
\cmidrule(lr){8-11}
\cmidrule(lr){14-18}
& \multicolumn{1}{c}{}
& \multicolumn{1}{c|}{}
& Pick Coke & Move Near & Drawer & Avg.
& Pick Coke & Move Near & Drawer & Avg.
& \multicolumn{1}{c}{}
& \multicolumn{1}{c|}{}
& Spoon & Carrot & Stack Cube & Eggplant & Avg. \\
\midrule
\rowcolor{gray!15}
Baseline (GR00T N1.7)
& 5.59 & 0.87
& 93.7 & 77.5 & 55.1 & 75.4
& 85.6 & 75.7 & 30.4 & 63.9
& 3.30 & 0.95
& 79.2 & 66.7 & 8.3 & 41.7 & 49.0 \\

\quad + TR
& 4.93 & 0.44
& 89.3 & \textbf{82.1} & 56.0 & 75.8
& 85.6 & 76.2 & 31.5 & 64.4
& 3.02 & 0.87
& 70.8 & \textbf{70.8} & \textbf{12.5} & 50.0 & 51.0 \\

\quad + AC
& 1.72 & 0.50
& 96.3 & 78.8 & 55.6 & 76.9
& 89.9 & 77.5 & \textbf{42.3} & \textbf{69.9}
& 1.67 & 0.46
& 75.0 & 54.2 & 0.0 & \textbf{83.3} & 53.1 \\

\quad + AC + TR (Ours)
& \textbf{1.53} & \textbf{0.39}
& \textbf{98.0} & 78.3 & \textbf{57.9} & \textbf{78.1}
& \textbf{90.4} & \textbf{77.7} & 40.7 & 69.6
& \textbf{1.49} & \textbf{0.21}
& \textbf{83.3} & 58.3 & 4.2 & 79.2 & \textbf{56.3} \\
\bottomrule
\end{tabular}%
}
\caption{\textbf{Evaluation on SimplerEnv.} Success rates (\%) on Google Robot and WidowX. VM and VA denote visual matching and variant aggregation, respectively. Metrics $M$ and $C$ measure action deviation magnitude and cross-condition directional consistency, respectively.}
\label{tab:simpler}
\vspace{-2mm}
\end{table*}

\subsection{Directional Bias Correction}

\begin{figure}[t]
  \centering
  \includegraphics[width=\linewidth, page=1]{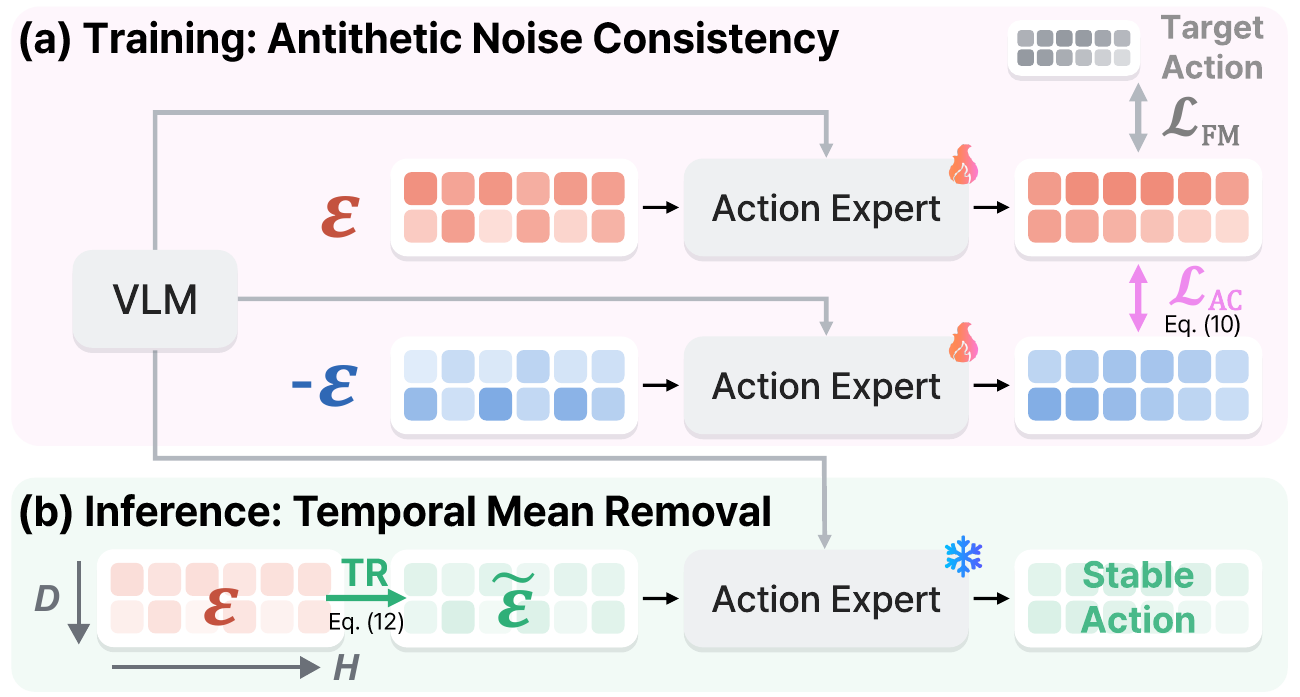}
  \caption{\textbf{Training and inference pipeline.} Antithetic noise consistency aligns denoised action estimates during training (a), while temporal mean removal centers the initial noise at inference (b).}
  \label{fig:fig_5}
  \vspace{-3mm}
\end{figure}

Our analysis highlights the need for methods that mitigate the directional bias associated with initial noise. One straightforward approach is to remove the temporal mean before the first denoising step. However, the action expert may already have learned to rely on this component as a directional prior during training. Removing this component can reduce directional bias, but noise sensitivity may persist because the learned noise-to-action mapping remains unchanged. We therefore train the action expert to suppress directional bias, as shown in~\cref{fig:fig_5}~(a), complementing temporal mean removal at inference in \cref{fig:fig_5}~(b).

Specifically, we introduce an \textbf{antithetic noise consistency~(AC)} term to encourage agreement between denoised action estimates obtained from antithetic noise pairs ($\epsilon$,$-\epsilon$).
During training, we construct interpolated inputs using this pair with the same target action sequence $x_0$ and timestep $t$.
For each branch, we compute a denoised action estimate $\widehat{x}_{0\mid t}=x_t-t\,v_\theta(x_t,t,c)$ from a single velocity evaluation, without running the full sampling process.
The antithetic noise consistency loss is defined as:
\begin{equation}
\mathcal L_{\mathrm{AC}}
=
\mathbb E\!\left[
\left\|
\widehat x_{0\mid t}(\epsilon)
-
\widehat x_{0\mid t}(-\epsilon)
\right\|_F^2
\right].
\label{eq:inverse_noise_consistency}
\end{equation}
We add this term to the original flow-matching objective in~\cref{eq:fm_training}, yielding the total loss:
\begin{equation}
\mathcal L
=
\mathcal L_{\mathrm{FM}}
+
\lambda\,\mathcal L_{\mathrm{AC}}.
\label{eq:inverse_noise_loss}
\end{equation}
Here, $\lambda\geq0$ controls the regularization strength. The flow-matching loss is evaluated on the $\epsilon$ branch, while gradients from the consistency term propagate through both branches.

\begin{table*}[t]
\centering
\resizebox{\textwidth}{!}{%
\begin{tabular}{l c|cccc|c|ccccccc|c|}
\toprule
\multirow{2}{*}{\vspace{-2mm}Method} & \multirow{2}{*}{\vspace{-2mm}\#Cand.} & \multicolumn{5}{c|}{LIBERO}
& \multicolumn{8}{c}{LIBERO-Plus} \\
\cmidrule(lr){3-7}
\cmidrule(lr){8-15}
& & Spatial & Object & Goal & Long & Avg. & Cam. & Robot & Lang. & Light & BG & Noise & Layout & Avg. \\
\midrule

\rowcolor{gray!15}
\multicolumn{15}{l}{\textit{VLAs (Base VLA)}} \\

OpenVLA & 1
& 84.7 & 88.4 & 79.2 & 53.7 & 76.5
& 0.8 & 3.5 & 23.0 & 8.1 & 34.8 & 15.2 & 28.5 & 15.6 \\

OpenVLA-OFT & 1
& 97.7 & 98.0 & 96.1 & 95.3 & 96.8
& 55.6 & 21.7 & 81.0 & 92.7 & 91.0
& \underline{78.6} & 68.7 & 67.9 \\

WorldVLA & 1
& 87.6 & 96.2 & 83.4 & 60.0 & 81.8
& 0.1 & 27.9 & 41.6 & 43.7 & 17.1 & 10.9 & 38.0 & 25.0 \\

UniVLA & 1
& 96.5 & 96.8 & 95.6 & 92.0 & 95.2
& 1.8 & 46.2 & 69.6 & 69.0 & 81.0 & 21.2 & 31.9 & 43.9 \\

X-VLA & 1
& 98.2 & 98.6 & 97.8 & \textbf{97.6} & \underline{98.1}
& 23.4 & \textbf{89.7} & 75.7 & 88.2 & \textbf{96.0}
& 62.7 & 71.8 & 71.4 \\

\midrule
\rowcolor{gray!15}
\multicolumn{15}{l}{\textit{Inference-Time Approaches} $^\ddagger$} \\

MGSelect & 4
& 97.2 & 98.0 & 94.5 & 82.7 & 93.1
& -- & -- & -- & -- & -- & -- & -- & -- \\

RoboMonkey & 16
& 98.6 & 99.0 & 97.6 & 94.0 & 97.3
& -- & -- & -- & -- & -- & -- & -- & -- \\

Hume & 5
& 98.6 & \textbf{99.8} & \textbf{99.4}
& \underline{96.7} & \textbf{98.6}
& -- & -- & -- & -- & -- & -- & -- & -- \\

TokenPB & 16
& 96.6 & 98.8 & 97.0 & 94.6 & 96.8
& 43.2 & 54.2 & \underline{88.4} & 93.6 & 92.7
& 74.6 & 79.7 & 73.5 \\

\midrule
\rowcolor{gray!15}
\multicolumn{15}{l}{\textit{Training-Time Approaches}} \\

BehaviorVLA & 1
& \underline{99.2} & \underline{99.4} & 98.8 & 94.6 & 98.0
& -- & -- & -- & -- & -- & -- & -- & -- \\

GeCO & 1
& 97.4 & 98.6 & 96.4 & 92.4 & 96.2
& -- & -- & -- & -- & -- & -- & -- & -- \\

ResVLA & 1
& 96.8 & 98.6 & 97.4 & 92.4 & 96.3
& 49.8 & 59.9 & \textbf{88.5} & 90.5 & \underline{94.9}
& 76.8 & 79.0 & 75.3 \\

\midrule
\rowcolor{green!15}
\multicolumn{15}{l}{\textit{Our Method} $^\dagger$} \\

GR00T N1.7 & 1
& 98.0 & 99.2 & 98.2 & 86.4 & 95.5
& \underline{57.3} & 61.8 & 82.3 & \underline{94.5} & 93.5
& 71.7 & \underline{80.9} & 75.8 \\

GR00T N1.7 + Ours & 1
& 99.0 & \underline{99.4} & 98.6 & 92.0 & 97.3
& \textbf{58.2} & 59.0 & 84.6 & \textbf{94.8} & 94.2
& \textbf{81.2} & 77.8 & \underline{77.0} \\

$\Delta$ &
& \textcolor{green!50!black}{$\boldsymbol{+1.0}$}
& \textcolor{green!50!black}{$\boldsymbol{+0.2}$}
& \textcolor{green!50!black}{$\boldsymbol{+0.4}$}
& \textcolor{green!50!black}{$\boldsymbol{+5.6}$}
& \textcolor{green!50!black}{$\boldsymbol{+1.8}$}
& \textcolor{green!50!black}{$\boldsymbol{+0.9}$}
& \textcolor{red!70!black}{$\boldsymbol{-2.8}$}
& \textcolor{green!50!black}{$\boldsymbol{+2.3}$}
& \textcolor{green!50!black}{$\boldsymbol{+0.3}$}
& \textcolor{green!50!black}{$\boldsymbol{+0.7}$}
& \textcolor{green!50!black}{$\boldsymbol{+9.5}$}
& \textcolor{red!70!black}{$\boldsymbol{-3.1}$}
& \textcolor{green!50!black}{$\boldsymbol{+1.2}$} \\

$\pi_{0.5}$ & 1
& 95.4 & 99.2 & 96.0 & 84.6 & 93.8
& 51.2 & 70.0 & 72.5 & 85.8 & 83.6
& 67.8 & 79.9 & 71.8 \\

$\pi_{0.5}$ + Ours & 1
& \textbf{99.4} & \textbf{99.8} & \underline{99.2}
& 96.0 & \textbf{98.6}
& 56.2 & \underline{75.7} & 77.0 & 91.7 & 89.5
& 73.8 & \textbf{85.6} & \textbf{77.3} \\

$\Delta$ &
& \textcolor{green!50!black}{$\boldsymbol{+4.0}$}
& \textcolor{green!50!black}{$\boldsymbol{+0.6}$}
& \textcolor{green!50!black}{$\boldsymbol{+3.2}$}
& \textcolor{green!50!black}{$\boldsymbol{+11.4}$}
& \textcolor{green!50!black}{$\boldsymbol{+4.8}$}
& \textcolor{green!50!black}{$\boldsymbol{+5.0}$}
& \textcolor{green!50!black}{$\boldsymbol{+5.7}$}
& \textcolor{green!50!black}{$\boldsymbol{+4.5}$}
& \textcolor{green!50!black}{$\boldsymbol{+5.9}$}
& \textcolor{green!50!black}{$\boldsymbol{+5.9}$}
& \textcolor{green!50!black}{$\boldsymbol{+6.0}$}
& \textcolor{green!50!black}{$\boldsymbol{+5.7}$}
& \textcolor{green!50!black}{$\boldsymbol{+5.5}$} \\

\bottomrule
\end{tabular}%
}
\caption{
\textbf{Success rates (\%) on LIBERO and LIBERO-Plus.}
\#Cand. denotes the number of action candidates per decision.
$^\dagger$ marks our measurements.
$^\ddagger$ indicates additional inference-time computation.
}
\label{tab:libero_results}
\vspace{-2mm}
\end{table*}

At inference, we apply \textbf{temporal mean removal (TR)} to the initial noise to reduce directional bias. For each action coordinate, we compute the mean across action steps and subtract it from every step:
\begin{equation}
\widetilde\epsilon
=
\epsilon-\mathbf1\bar\epsilon^\top,
\quad s.t.\quad
\bar\epsilon
=
\frac1H\epsilon^\top\mathbf1.
\label{eq:temporal_mean_removal}
\end{equation}
Finally, the action expert generates actions from the mean-centered noise ($\widetilde{\epsilon}$) without additional model evaluations.

\section{Experiments}
\subsection{Simulation Experiments}
\noindent\textbf{Implementation details.}\quad
\label{subsec:implementation}
We implement our method using pretrained \(\pi_{0.5}\) and GR00T N1.7 as backbone models.
We fine-tune $\pi_{0.5}$ using LoRA~\cite{hu2021lora} with rank 16 for the vision-language backbone and rank 32 for the action expert. In both modules, we apply LoRA to the attention and feed-forward layers and set $\alpha$ equal to the rank.
For GR00T N1.7, we fine-tune the projector and diffusion transformer while keeping the vision and language encoders frozen.
We use the Fractal~\cite{brohan2022rt}, Bridge V2~\cite{walke2023bridgedata}, and LIBERO~\cite{liu2023libero} datasets for training.
We set $\lambda=3$ for $\pi_{0.5}$ and $\lambda=5$ for GR00T N1.7.
We optimize both models with AdamW~\cite{loshchilov2017decoupled} using a learning rate of $10^{-4}$.
For SimplerEnv~\cite{li2024evaluating}, we fine-tune GR00T N1.7 on Fractal and Bridge V2 for 40k and 20k steps, using an effective batch size of 1{,}024 for both datasets.
On LIBERO, we train both $\pi_{0.5}$ and GR00T N1.7 for 30k steps, with batch sizes of 32 and 64.
Training on Fractal and Bridge V2 takes 30 hours using four NVIDIA RTX PRO 6000 Blackwell GPUs, while LIBERO training takes 40 hours on a single GPU.

\label{subsec:simulation}
\noindent\textbf{Evaluation setup.}\quad
We evaluate our method on SimplerEnv, LIBERO, and LIBERO-Plus~\cite{fei2025libero}.
SimplerEnv includes three Google Robot tasks evaluated under visual matching (VM) and variant aggregation (VA), as well as four WidowX tasks.
For SimplerEnv, we use the official GR00T evaluation settings, with 756 VM and 1,803 VA episodes for Google Robot and 24 initial configurations per WidowX task.
LIBERO comprises 40 tasks across four suites, with 50 initial states per task, for a total of 2,000 episodes. We further evaluate LIBERO-trained policies on all 10,030 LIBERO-Plus instances.

\begin{figure}[t]
  \centering
   \includegraphics[width=1\linewidth]{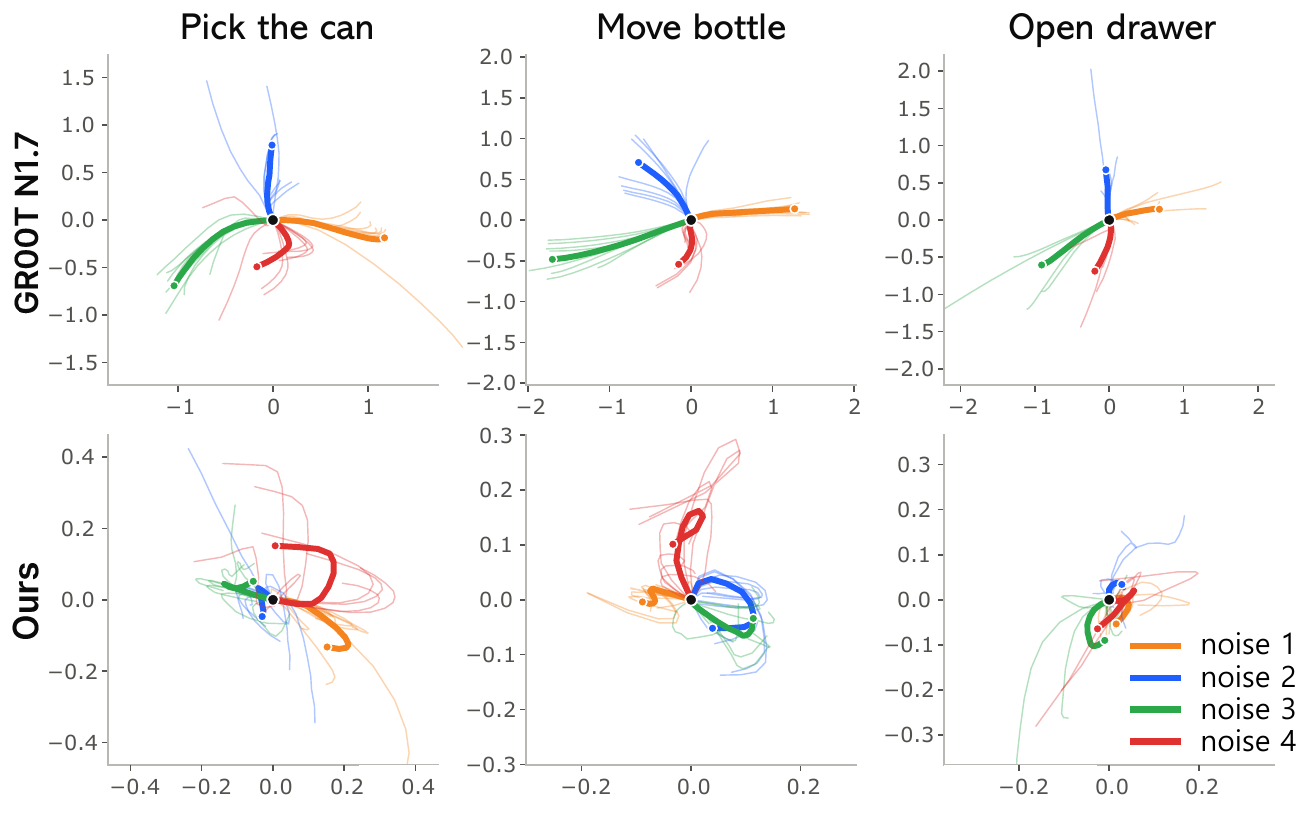}
    \caption{\textbf{Action deviations before and after bias correction.} The baseline and our method are shown with different axis ranges.}
   \label{fig:fig_6}
   \vspace{-3mm}
\end{figure}

\noindent\textbf{Noise sensitivity and task performance.}\quad
As shown in~\cref{tab:simpler}, we apply the two corrections to GR00T N1.7 and evaluate their effects on SimplerEnv. We report the deviation magnitude ($M$) and cross-condition directional consistency ($C$) to assess noise sensitivity and task performance for robust execution.
We define $M$ as:
\begin{equation}
M = \mathbb{E}_{c,\epsilon}
\left[
\left\|y_0(\epsilon,c)-\overline{y}_0(c)\right\|_F^2
\right],
\label{eq:noise_induced_variance}
\end{equation}
where $\overline{y}_0(c)=\mathbb{E}_{\epsilon}[y_0(\epsilon,c)]$ is the noise-averaged action sequence.\,This metric captures the magnitude of action deviations associated with directional bias.\,We also define $C$ as:
\begin{equation}
C = \mathbb{E}_{\epsilon}\mathbb{E}_{c\neq c'}
\left[
\cos\!\left(b_c(\epsilon),b_{c'}(\epsilon)\right)
\right].
\label{eq:cross_condition_consistency}
\end{equation}
This metric captures the directional consistency of the bias induced by the same noise sample across different conditions. We compute both metrics using 120 observations from each of Fractal and Bridge and the same 64 base noise samples per observation across variants.

We compare TR alone, AC alone, and their combination (AC+TR) against the baseline in terms of $M$, $C$, and task success rate.
As intended, both methods reduce $M$ and $C$ relative to the baseline, with their combination achieving the lowest values on both datasets.
Our method consistently improves success rates, achieving absolute gains of 2.7\% under visual matching (VM) and 5.7\% under variant aggregation (VA) on Google Robot, and 7.3\% on WidowX.
These results mean the task performance can be improved simply by reducing directional bias across robot embodiments and evaluation settings.

\begin{figure*}[t]
\centering
\includegraphics[width=\linewidth]{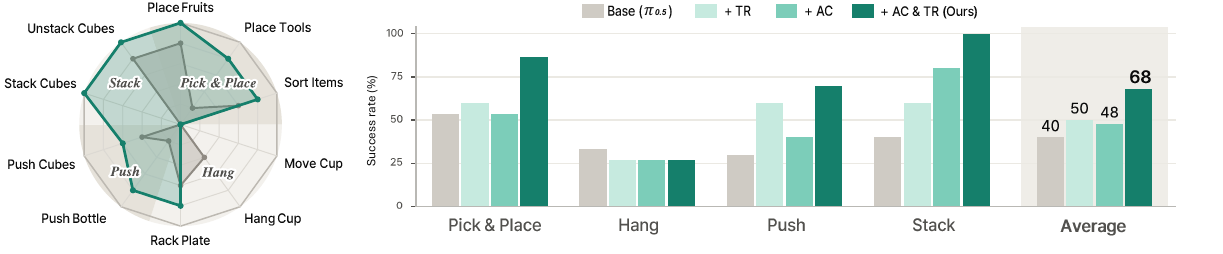}
\caption{\textbf{Real-robot success rates (\%).} We evaluate four $\pi_{0.5}$ variants across 10 manipulation tasks with five trials per task.}
\label{fig:fig_7}
\vspace{-3mm}
\end{figure*}

\noindent\textbf{Visualization of directional bias.}\quad
We compare action deviations between the baseline and our model in~\cref{fig:fig_6}.
Consistent with the observations in~\cref{subsec:empirical_observations}, the baseline exhibits a similar directional bias across tasks for the same noise sample. In contrast, our model produces less consistent bias directions across tasks, reflecting the reduction in \(C\). The deviations also decrease in magnitude, consistent with a lower \(M\), as reflected in the smaller axis ranges in the bottom row. Together, these observations suggest that our corrections reduce action deviations while weakening the directional influence of the initial noise’s temporal mean.

\noindent\textbf{Quantitative comparison.}\quad
In~\cref{tab:libero_results}, we compare our method with base VLA policies and existing inference-time and training-time approaches on LIBERO and LIBERO-Plus. Our method improves the average success rate of both models on both benchmarks. For \(\pi_{0.5}\), the average success rate increases from 93.8\% to 98.6\% on LIBERO and from 71.8\% to 77.3\% on LIBERO-Plus. For GR00T N1.7, average success increases from 95.5\% to 97.3\% on LIBERO and from 75.8\% to 77.0\% on LIBERO-Plus. Across the perturbation categories in LIBERO-Plus, \(\pi_{0.5}\) improves in all seven categories, while GR00T N1.7 improves in five, with declines limited to Robot and Layout. These gains across diverse perturbations suggest improved generalization to perturbed environments. Among the compared methods, our \(\pi_{0.5}\) variant matches the highest average success rate on LIBERO and achieves the highest on LIBERO-Plus. Hume achieves the same LIBERO success rate but evaluates five action candidates per decision. In contrast, our method achieves this performance with a single candidate, avoiding additional inference-time overhead.

\begin{figure}[t]
  \centering
   \includegraphics[width=1\linewidth]{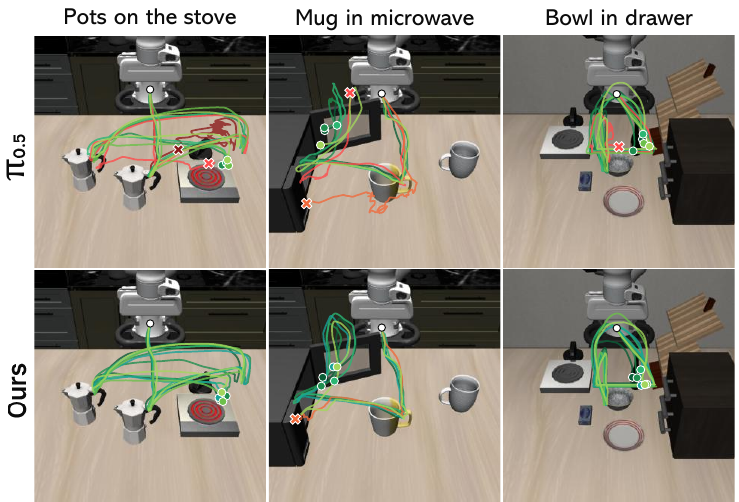}
    \caption{\textbf{LIBERO execution trajectories.} Green and red denote successful and failed trials for the baseline and our method.}
   \label{fig:fig_8}
\end{figure}

\noindent\textbf{Qualitative Results.}\quad
In~\cref{fig:fig_8}, we report end-effector trajectories of $\pi_{0.5}$ in LIBERO simulations, with green denoting success and red denoting failure. Our method produces highly consistent paths across successful trials and exhibits limited deviations even when a trial fails. The baseline, however, frequently moves far from the intended path. These departures can expose the policy to out-of-distribution observations, making it difficult to recover after an inaccurate action. These observations show that directional bias can affect the entire rollout and substantially reduce execution reliability.

\begin{figure}[t]
\centering
\includegraphics[width=\linewidth]{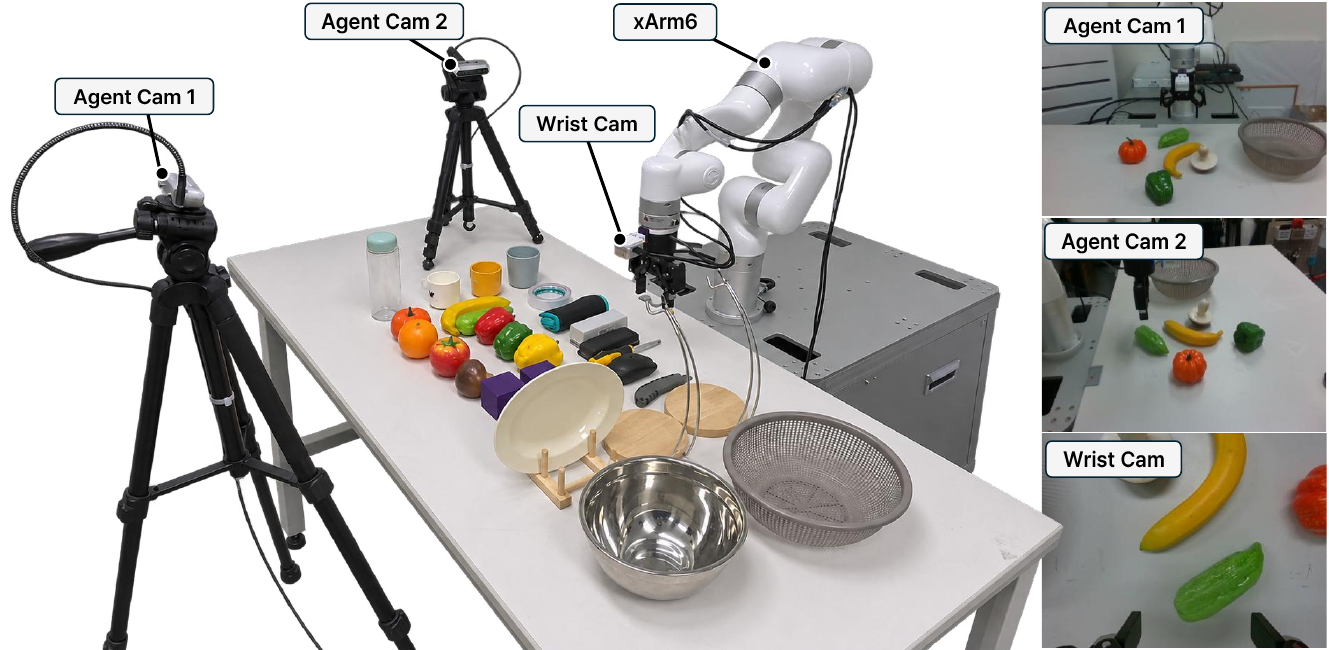}
\caption{\textbf{Real-robot setup.} An xArm6 with a Robotiq 2F-85 and three cameras is shown alongside example camera observations.}
\label{fig:fig_9}
\vspace{-2mm}
\end{figure}

\subsection{Real-Robot Experiments}
\label{sec:real_exp}

\noindent\textbf{Experimental setup.}\quad
We further deploy \(\pi_{0.5}\) on a real robot to examine whether the improvements observed in simulation extend to physical execution.
As shown in~\cref{fig:fig_9},
we use a UFACTORY xArm6 robotic arm equipped with a Robotiq 2F-85 gripper, two RealSense D435 cameras for external views, and a RealSense D405 wrist camera.
We evaluate 10 manipulation tasks across four categories: Pick \& Place, Hang, Push, and Stack.
For each task, we collect 50 demonstrations through teleoperation~\cite{welle2024quest2ros} using a Meta Quest Pro headset.
We fine-tune the baseline and our model on the same collected demonstrations for 30k steps with a batch size of 32, using the same LoRA configuration as on LIBERO.
The four variants are the baseline, AC alone, TR alone, and AC+TR. We conduct five trials per task for each variant and report the overall success rate across all 10 tasks.

\noindent\textbf{Comparison of task performance.}\quad
As shown in~\cref{fig:fig_7}, our method achieves the highest average success rate of 68\%, compared with 40\% for the baseline.
Our approach improves performance in all categories except Hang, suggesting that noise correction alone may not compensate for insufficiently learned task behaviors. However, in the other three categories, all methods match or exceed baseline performance, with our method achieving the highest success rate in each category. During execution, we also observe that directional bias can make it difficult for the robot to recover and continue the task. In particular, the baseline model frequently produces downward motions that press into objects and cause the robot to stop. These failures are substantially reduced by our corrections and disappear entirely when AC and TR are combined. This shows that our method helps the policy to execute behaviors more reliably, supporting more robust and safer real-world deployment.

\subsection{Ablation Studies}

\begin{figure}[t]
  \centering
  \includegraphics[width=\linewidth]{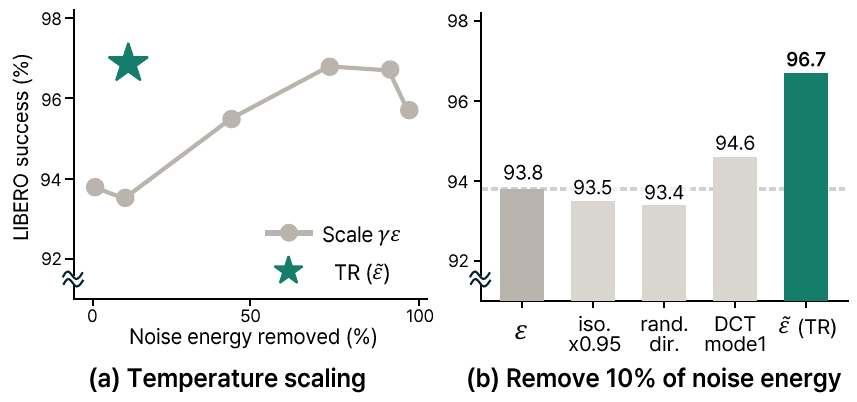}
    \caption{\textbf{Noise energy and performance.}
    LIBERO success rates (\%) of the $\pi_{0.5}$ baseline.
    (a) Isotropic scaling $\gamma\epsilon$ for $\gamma \in \{1, 0.95, 0.75, 0.5, 0.25, 0\}$, compared with TR.
    (b) Removing $10\%$ of the noise energy along different directions.}
  \label{fig:fig_10}

\end{figure}

\noindent\textbf{Noise energy and performance.}\quad
To understand the source of TR's improvement, we compare the effects of removing the temporal mean component and reducing noise energy.
We define the energy of noise correction:
\begin{equation}
E(z)=\|z\|_F^2, \quad E(\gamma\epsilon)=\gamma^2 E(\epsilon),
\label{eq:energy}
\end{equation}
where $z$ is the noise initialization and $\gamma$ is the isotropic scaling factor.
We first reduce noise energy by scaling and report LIBERO success rates in \cref{fig:fig_10}~(a).
However, the success rate does not exceed the performance of TR, whereas removing all noise ($z=0$) lowers it to $95.7\%$. We then compare TR with three alternatives at approximately matched energy reduction: isotropic scaling ($0.95\epsilon$), removal of the component along a random temporal direction orthogonal to $\mathbf{1}$, and removal of the lowest non-constant DCT mode.
As shown in \cref{fig:fig_10}~(b), TR achieves the highest success rate among these methods at $96.7\%$. Together, these results suggest that selectively removing the temporal mean component contributes to TR's gains beyond the effect of reducing total noise energy.

\noindent\textbf{Alternative training objectives.}\quad
In~\cref{tab:ablation_objectives}, we compare baseline and consistency training with different noise configurations.
Baseline training applies the original objective to either one noise sample $\epsilon$ or two independent samples $(\epsilon,\epsilon')$.
For consistency training, we additionally replace antithetic pairs $(\epsilon,-\epsilon)$ with independent pairs $(\epsilon,\epsilon')$ in AC to assess the benefit of antithetic pairing.
Using two samples for baseline training does not improve performance, suggesting that AC's gains cannot be explained by the additional sample alone. With consistency training, antithetic pairs outperform independent pairs on both benchmarks, increasing success rates from 94.9\% to 96.9\% on LIBERO and from 70.7\% to 75.6\% on LIBERO-Plus. These results highlight the benefit of targeting directional bias through antithetic noise pairing, beyond simply encouraging consistency across different noise samples.

\noindent\textbf{Effect of $\lambda$.}\quad
\cref{tab:ablation_beta} shows that $\lambda=3$ performs best both with and without temporal mean removal, reaching 98.6\% on LIBERO and 77.3\% on LIBERO-Plus with temporal mean removal. Increasing $\lambda$ to 10 or 30 reduces performance on both benchmarks, possibly because an overly strong consistency constraint interferes with learning to match the ground-truth actions. We therefore use $\lambda=3$ for the main $\pi_{0.5}$ LIBERO experiments.

\begin{table}[t]
\centering

\small
\setlength{\tabcolsep}{4pt}
\resizebox{\columnwidth}{!}{%
\begin{tabular}{@{}lcccc@{}}
\toprule
& \multicolumn{2}{c}{Without TR}%
& \multicolumn{2}{c}{With TR}\\%
\cmidrule(lr){2-3}
\cmidrule(lr){4-5}
Training objective
& LIBERO & LIBERO-Plus
& LIBERO & LIBERO-Plus \\
\midrule
Baseline $(\boldsymbol{\epsilon})$
& 93.8 & \underline{71.8}
& \underline{96.7} & \underline{75.3} \\
Baseline $(\boldsymbol{\epsilon},\boldsymbol{\epsilon}')$
& 93.8 & 70.1
& 95.4 & 74.1 \\
AC $(\boldsymbol{\epsilon},\boldsymbol{\epsilon}')$
& \underline{94.9} & 70.7
& \underline{96.7} &  72.1 \\
\midrule
AC
& \textbf{96.9} & \textbf{75.6}
& \textbf{98.6} & \textbf{77.3} \\
\bottomrule
\end{tabular}%
}
\caption{\textbf{Training objective ablation.} Success rates (\%) on LIBERO and LIBERO-Plus.}
\label{tab:ablation_objectives}
\end{table}

\begin{table}[t]
\centering

\small
\setlength{\tabcolsep}{4pt}
\resizebox{0.8\columnwidth}{!}{%
\begin{tabular}{@{}ccccc@{}}
\toprule
& \multicolumn{2}{c}{Without TR}
& \multicolumn{2}{c}{With TR} \\
\cmidrule(lr){2-3}
\cmidrule(lr){4-5}
$\lambda$
& LIBERO & LIBERO-Plus
& LIBERO & LIBERO-Plus \\
\midrule
0
& 93.8 & 71.8
& 96.7 & \underline{75.3} \\
1
& 94.1 & 72.2
& \underline{97.2} & 75.1 \\
3
& \textbf{96.9} & \textbf{75.6}
& \textbf{98.6} & \textbf{77.3} \\
10
& \underline{96.1} & \underline{73.7}
& 96.8 & 74.1 \\
30
& 96.0 & 69.2
& 96.5 & 68.8 \\
\bottomrule
\end{tabular}%
}
\caption{\textbf{Regularization weight ablation.} Success rates (\%) on LIBERO and LIBERO-Plus.}
\label{tab:ablation_beta}
\end{table}

\section{Conclusion}
We have presented a simple framework for improving the robustness of vision-language-action models to initial noise. We identify a systematic directional bias in generated actions that closely aligns with the noise's temporal mean and explain this alignment in conditional expectation under a Gaussian action model. Based on these findings, we combine antithetic noise consistency during training with temporal mean removal at inference to reduce this bias. Experiments on SimplerEnv, LIBERO, and LIBERO-Plus demonstrate reduced noise sensitivity and improved task performance. Furthermore, real-robot experiments validate the benefits of these corrections for physical execution. Our results highlight the importance of correcting directional bias for reliable deployment of VLA policies.

\noindent\textbf{Limitations.}\quad
Our theoretical analysis relies on Gaussian conditional action distributions and ideal flow matching, which may not fully capture the behavior of learned policies in real-world settings.
While our method introduces no additional inference-time overhead, antithetic noise consistency requires additional training time to evaluate paired noise inputs.
However, encouraging consistent predictions may also reduce useful action diversity when multiple solutions are valid. Removing directional bias while preserving such diversity remains an open challenge.

{
    \small
    \bibliographystyle{ieeenat_fullname}
    \bibliography{references}
}
\maketitlesupplementary
\appendix
\makeatletter
\setlength{\@fptop}{0pt}
\makeatother

\noindent\textbf{Overview.}
The main paper relates the temporal mean of the initial noise to directional bias in conditional expectation, and shows that TR at inference and AC during training improve task success.
In this supplement, we address three questions that follow from these results.
We first prove the closed-form generation of Eq.~\eqref{eq:closed_form_generation} in \cref{app:closed_form_generation}.
We then show that the conditional-expectation analysis of Sec.~3.3 also describes the directional bias of individual noise samples in \cref{app:additional_theory}.
We examine whether the gain of TR comes from removing the temporal mean rather than from reducing the noise energy in \cref{app:remove}.
We extend this account of directional bias beyond the idealized setting, in theory in \cref{app:robustness} and on a second backbone deployed on a real robot in \cref{app:real_robot}.
Finally, we provide implementation details in \cref{app:impl}.

\section{Proof of Closed-Form Action Generation}
\label{app:closed_form_generation}

We provide a detailed derivation of Eq.~\eqref{eq:closed_form_generation} under the Gaussian action model in Eq.~\eqref{eq:action_gaussian}. Throughout this section, the conditioning input $c$ is fixed, and all expectations and covariances are conditional on $c$ unless stated otherwise. The target action sequence $x_0\in\mathbb{R}^{H\times d}$ has independent action dimensions, corresponding to its columns, with conditional mean $\mu_c$ and a shared temporal covariance $\Sigma_c=\Sigma_c^\top\succ0$:
\begin{equation}
(x_0)_{:,j}\mid c
\sim\mathcal{N}\!\left((\mu_c)_{:,j},\Sigma_c\right),
\qquad j=1,\ldots,d.
\label{eq:app_action_distribution}
\end{equation}
The initial noise $\epsilon\in\mathbb{R}^{H\times d}$ contains independent standard Gaussian entries and is independent of $(x_0,c)$. The derivation applies to any symmetric positive-definite $\Sigma_c$. All matrix square roots below are principal symmetric positive-definite square roots.

\paragraph{Population-optimal velocity field.}
Consider the linear training interpolation and its target velocity:
\begin{equation}
x_t=(1-t)x_0+t\epsilon,
\qquad
u=\epsilon-x_0.
\label{eq:app_training_interpolation}
\end{equation}
For a fixed $t$, squared-error regression is minimized by the conditional expectation of the target. We therefore define
\begin{equation}
v_\star(x,t,c)
=
\mathbb{E}[u\mid x_t=x,c].
\label{eq:app_conditional_velocity}
\end{equation}
Indeed, for any candidate velocity field $v$ with finite squared error, the conditional-mean property gives the orthogonal decomposition
\begin{equation}
\begin{aligned}
&\mathbb{E}\!\left[
\|v(x_t,t,c)-u\|_F^2
\right]\\
&=
\mathbb{E}\!\left[
\|v(x_t,t,c)-v_\star(x_t,t,c)\|_F^2
\right]\\
&\quad+
\mathbb{E}\!\left[
\|v_\star(x_t,t,c)-u\|_F^2
\right].
\end{aligned}
\label{eq:app_squared_error_decomposition}
\end{equation}
The second term does not depend on $v$, while the first is minimized when $v=v_\star$ almost surely. Averaging this identity over the training-time distribution of $t$ establishes population optimality for the flow-matching objective. Below, we use the continuous conditional-mean velocity field obtained explicitly from the Gaussian model.

\paragraph{Gaussian conditioning.}
Since $x_0$ and $\epsilon$ are independent and Gaussian at fixed $c$, the target velocity $u$ and interpolation $x_t$ are jointly Gaussian. Their means are
\begin{equation}
\mathbb{E}[x_t]=(1-t)\mu_c,
\qquad
\mathbb{E}[u]=-\mu_c.
\label{eq:app_interpolation_means}
\end{equation}
For each action dimension $j$, independence eliminates the cross-covariance terms and gives
\begin{equation}
\begin{aligned}
C_t
&:=\operatorname{Cov}\!\left((x_t)_{:,j}\right)\\
&=(1-t)^2\operatorname{Cov}\!\left((x_0)_{:,j}\right)
+t^2\operatorname{Cov}(\epsilon_{:,j})\\
&=(1-t)^2\Sigma_c+t^2I,
\end{aligned}
\label{eq:app_interpolation_covariance}
\end{equation}
where $I$ is the $H\times H$ identity matrix. Similarly, the cross-covariance between the target velocity and the interpolation is
\begin{equation}
\begin{aligned}
B_t
&:=\operatorname{Cov}\!\left(u_{:,j},(x_t)_{:,j}\right)\\
&=t\operatorname{Cov}(\epsilon_{:,j})
-(1-t)\operatorname{Cov}\!\left((x_0)_{:,j}\right)\\
&=tI-(1-t)\Sigma_c.
\end{aligned}
\label{eq:app_velocity_cross_covariance}
\end{equation}
We suppress the dependence of $C_t$ and $B_t$ on $c$ for readability. The matrix $C_t$ is positive definite for every $t\in[0,1]$, so its inverse exists. Applying the Gaussian conditional-mean formula yields
\begin{equation}
\begin{aligned}
&\mathbb{E}\!\left[u_{:,j}\mid (x_t)_{:,j}=z,c\right]\\
&=-(\mu_c)_{:,j}
+B_tC_t^{-1}\bigl[z-(1-t)(\mu_c)_{:,j}\bigr].
\end{aligned}
\label{eq:app_coordinate_conditioning}
\end{equation}
Because the action dimensions are independent, these conditional means combine into the matrix-valued velocity field
\begin{equation}
v_\star(x,t,c)
=
-\mu_c+B_tC_t^{-1}\bigl[x-(1-t)\mu_c\bigr].
\label{eq:app_optimal_velocity}
\end{equation}

\paragraph{Differentiating the covariance square root.}
To solve the sampling ODE, we first establish the required derivatives of $C_t^{1/2}$ and $C_t^{-1/2}$. Let
\begin{equation}
\begin{aligned}
\Sigma_c
&=Q\operatorname{diag}(\sigma_1,\ldots,\sigma_H)Q^\top,\\
Q^\top Q&=I,\qquad \sigma_i>0.
\end{aligned}
\label{eq:app_covariance_eigendecomposition}
\end{equation}
Since $C_t$ and $B_t$ are functions of $\Sigma_c$, they share the same eigenvectors:
\begin{equation}
\begin{aligned}
C_t
&=Q\operatorname{diag}(c_1(t),\ldots,c_H(t))Q^\top,\\
B_t
&=Q\operatorname{diag}(b_1(t),\ldots,b_H(t))Q^\top,
\end{aligned}
\label{eq:app_shared_eigenbasis}
\end{equation}
where
\begin{equation}
\begin{aligned}
c_i(t)&=(1-t)^2\sigma_i+t^2,\\
b_i(t)&=t-(1-t)\sigma_i.
\end{aligned}
\label{eq:app_covariance_eigenvalues}
\end{equation}
Each $c_i(t)$ is strictly positive, and
$\frac{\mathrm{d}c_i(t)}{\mathrm{d}t}=2b_i(t)$.
Differentiating in this fixed eigenbasis therefore gives
\begin{equation}
\begin{aligned}
\frac{\mathrm{d}C_t^{1/2}}{\mathrm{d}t}
&=
Q\operatorname{diag}\!\left(
\frac{b_i(t)}{\sqrt{c_i(t)}}
\right)Q^\top\\
&=B_tC_t^{-1/2},\\
\frac{\mathrm{d}C_t^{-1/2}}{\mathrm{d}t}
&=
-Q\operatorname{diag}\!\left(
\frac{b_i(t)}{c_i(t)^{3/2}}
\right)Q^\top\\
&=-B_tC_t^{-3/2}.
\end{aligned}
\label{eq:app_covariance_root_derivatives}
\end{equation}
These identities hold because $B_t$, $C_t$, and their matrix powers commute.

\paragraph{Solving the sampling ODE.}
The ideal sampling trajectory satisfies
\begin{equation}
\frac{\mathrm{d}y_t^\star}{\mathrm{d}t}
=
v_\star(y_t^\star,t,c),
\qquad
y_1^\star=\epsilon,
\label{eq:app_sampling_ode}
\end{equation}
and is integrated from $t=1$ to $t=0$. Define the centered trajectory
\begin{equation}
W_t:=y_t^\star-(1-t)\mu_c.
\label{eq:app_centered_trajectory}
\end{equation}
Substituting Eq.~\eqref{eq:app_optimal_velocity} into Eq.~\eqref{eq:app_sampling_ode} gives
\begin{equation}
\frac{\mathrm{d}W_t}{\mathrm{d}t}
=
\frac{\mathrm{d}y_t^\star}{\mathrm{d}t}+\mu_c
=
B_tC_t^{-1}W_t.
\label{eq:app_centered_ode}
\end{equation}
Using Eq.~\eqref{eq:app_covariance_root_derivatives}, we obtain
\begin{equation}
\begin{aligned}
\frac{\mathrm{d}}{\mathrm{d}t}
\left(C_t^{-1/2}W_t\right)
&=-B_tC_t^{-3/2}W_t\\
&\quad+C_t^{-1/2}B_tC_t^{-1}W_t\\
&=0.
\end{aligned}
\label{eq:app_normalized_trajectory_invariant}
\end{equation}
Thus, $C_t^{-1/2}W_t$ remains constant along the sampling trajectory. At $t=1$, we have $C_1=I$ and $W_1=y_1^\star=\epsilon$, so
\begin{equation}
C_t^{-1/2}W_t=\epsilon.
\label{eq:app_normalized_trajectory}
\end{equation}
Multiplying by $C_t^{1/2}$ and restoring the mean yields
\begin{equation}
y_t^\star
=
(1-t)\mu_c+C_t^{1/2}\epsilon.
\label{eq:app_gaussian_flow_trajectory}
\end{equation}
This expression also directly verifies the ODE:
\begin{equation}
\begin{aligned}
\frac{\mathrm{d}y_t^\star}{\mathrm{d}t}
&=-\mu_c+B_tC_t^{-1/2}\epsilon\\
&=-\mu_c+B_tC_t^{-1}
\bigl[y_t^\star-(1-t)\mu_c\bigr]\\
&=v_\star(y_t^\star,t,c).
\end{aligned}
\label{eq:app_ode_verification}
\end{equation}
The coefficient $B_tC_t^{-1}$ is continuous and bounded on $[0,1]$, making the affine velocity field globally Lipschitz in its state, uniformly over $t$. The ODE therefore has a unique solution for the specified initial condition. Finally, $C_0=\Sigma_c$ gives
\begin{equation}
y_0^\star
=
\mu_c+\Sigma_c^{1/2}\epsilon,
\label{eq:app_closed_form_generation}
\end{equation}
which proves Eq.~\eqref{eq:closed_form_generation}.

\paragraph{Marginal distributions and conditional mean.}
For each action dimension, Eq.~\eqref{eq:app_gaussian_flow_trajectory} implies
\begin{equation}
(y_t^\star)_{:,j}\mid c
\sim
\mathcal{N}\!\left((1-t)(\mu_c)_{:,j},C_t\right).
\label{eq:app_sampling_marginals}
\end{equation}
These are exactly the marginal distributions of the training interpolation $x_t$. However, matching marginals does not imply matching individual trajectories: $y_t^\star$ is determined by the initial noise $\epsilon$ at fixed $c$, whereas $x_t$ is constructed from an independently sampled pair $(x_0,\epsilon)$. In particular, the generated endpoint $y_0^\star$ need not equal the training target $x_0$ associated with that pair.

Since $\epsilon$ is independent of $c$ and has zero mean, the ideal generated conditional mean is
\begin{equation}
\begin{aligned}
\mathbb{E}[y_0^\star\mid c]
&=\mu_c+\Sigma_c^{1/2}\mathbb{E}[\epsilon\mid c]\\
&=\mu_c.
\end{aligned}
\label{eq:app_generated_conditional_mean}
\end{equation}
The empirical reference in \cref{subsec:simulation} instead estimates the learned policy's conditional mean from finitely many generated samples.

\section{Per-Sample Directional Bias}
\label{app:additional_theory}

Sec.~3.3 shows that the directional bias is proportional to the temporal mean of the initial noise in conditional expectation (Eq.~\eqref{eq:conditional_mean_bias}).
In this section, we examine whether this relationship also holds for individual noise samples.
Under the ideal Gaussian model, we show that the directional bias of each sample splits exactly into a term proportional to the temporal mean and a residual.
We then measure this decomposition in all evaluated models without averaging over conditions.
Throughout, we use the Gaussian action model of Eq.~\eqref{eq:action_gaussian} and the closed-form generation of Eq.~\eqref{eq:closed_form_generation}.

\paragraph{Noise decomposition.}
We first split the initial noise into its temporal mean and the remaining mean-centered part.
Let $P=I-\tfrac{1}{H}\mathbf{1}\mathbf{1}^{\top}$ denote the projection that removes the temporal mean.
The mean-centered noise of Eq.~\eqref{eq:temporal_mean_removal} is $\widetilde{\epsilon}=P\epsilon$, which gives
\begin{equation}
\epsilon=\mathbf{1}\bar{\epsilon}^{\top}+\widetilde{\epsilon}.
\label{eq:app_noise_split}
\end{equation}
Each column of $\epsilon$ is a standard Gaussian vector, and its two parts are jointly Gaussian with zero cross-covariance, since $P\mathbf{1}=0$.
The temporal mean and the mean-centered noise are therefore independent, and the mean-centered noise has zero mean.
Since the noise is sampled independently of $c$, both properties also hold given $c$.
Taking the conditional expectation of Eq.~\eqref{eq:app_noise_split} then gives
\begin{equation}
\mathbb{E}[\epsilon\mid\bar{\epsilon},c]=\mathbf{1}\bar{\epsilon}^{\top},
\label{eq:app_noise_mean}
\end{equation}
which proves Eq.~\eqref{eq:conditional_noise_mean}.

\paragraph{Gain and residual direction.}
The temporal covariance enters the directional bias only through the vector $\Sigma_c^{1/2}\mathbf{1}$ in Eq.~\eqref{eq:generated_temporal_bias}.
We split this vector into a component along $\mathbf{1}$ and a component orthogonal to it:
\begin{equation}
\begin{aligned}
&\Sigma_c^{1/2}\mathbf{1}=g_c\mathbf{1}+r_c,\\
&g_c=\frac{1}{H}\mathbf{1}^{\top}\Sigma_c^{1/2}\mathbf{1},
\qquad
r_c=P\,\Sigma_c^{1/2}\mathbf{1}.
\end{aligned}
\label{eq:app_gain_residual}
\end{equation}
We refer to $g_c$ as the gain and to $r_c$ as the residual direction.
The gain is the scale that appears in Eq.~\eqref{eq:conditional_mean_bias}.
Since $\Sigma_c^{1/2}$ is positive definite, the gain lies between the square roots of the smallest and largest eigenvalues of $\Sigma_c$, and is therefore positive for every condition.

\paragraph{Per-sample decomposition.}
Substituting Eqs.~\eqref{eq:app_noise_split} and~\eqref{eq:app_gain_residual} into Eq.~\eqref{eq:generated_temporal_bias} gives
\begin{equation}
b_c(\epsilon)=g_c\,\bar{\epsilon}+\eta_c(\epsilon),
\qquad
\eta_c(\epsilon)=\frac{1}{H}\,\widetilde{\epsilon}^{\top}r_c,
\label{eq:app_per_sample}
\end{equation}
where the cross terms vanish because $\widetilde{\epsilon}^{\top}\mathbf{1}=0$ and $\mathbf{1}^{\top}r_c=0$.
The first term is the conditional expectation of Eq.~\eqref{eq:conditional_mean_bias}, which now appears in every sample rather than only on average.
The second term is a residual that depends only on the mean-centered noise.
It is therefore independent of $\bar{\epsilon}$ and has zero mean.
Each coordinate of the residual has variance $\|r_c\|_2^2/H^2$, whereas each coordinate of the aligned term has variance $g_c^2/H$.

\paragraph{Alignment with the temporal mean.}
Because the two terms are independent, the correlation $\rho_c$ between a coordinate of $b_c(\epsilon)$ and the corresponding coordinate of $\bar{\epsilon}$ is determined by their variances:
\begin{equation}
\rho_c^2
=\frac{g_c^2/H}{g_c^2/H+\|r_c\|_2^2/H^2}
=\frac{\bigl(\mathbf{1}^{\top}\Sigma_c^{1/2}\mathbf{1}\bigr)^2}{H\,\mathbf{1}^{\top}\Sigma_c\mathbf{1}}
\in(0,1].
\label{eq:app_rho}
\end{equation}
The second expression follows from $g_c^2H+\|r_c\|_2^2=\|\Sigma_c^{1/2}\mathbf{1}\|_2^2=\mathbf{1}^{\top}\Sigma_c\mathbf{1}$, and the upper bound follows from the Cauchy--Schwarz inequality.
The correlation equals one, so that $b_c(\epsilon)=g_c\bar{\epsilon}$ for every sample, exactly when the residual direction vanishes, that is, when $\mathbf{1}$ is an eigenvector of $\Sigma_c$.
Since each coordinate pair is a centered bivariate Gaussian, the bias and the temporal mean also share the same sign with probability $\tfrac{1}{2}+\tfrac{1}{\pi}\arcsin\rho_c$, which exceeds one half for every condition.

\paragraph{Noise negation.}
Negating the initial noise negates both terms of Eq.~\eqref{eq:app_per_sample}, so $b_c(-\epsilon)=-b_c(\epsilon)$ for every sample, including the residual.
This is the per-sample counterpart of Observation~2 under the ideal Gaussian model.

\begin{table*}[t]
\centering
\small
\setlength{\tabcolsep}{4.5pt}
\caption{\textbf{Per-sample statistics of directional bias.} Statistics are computed for individual samples without averaging over conditions.}
\label{tab:per_sample_models}
\begin{tabular}{@{}llcc ccc ccc ccc ccc@{}}
\toprule
& & & & \multicolumn{3}{c}{Baseline} & \multicolumn{3}{c}{+ TR} & \multicolumn{3}{c}{+ AC} & \multicolumn{3}{c}{+ AC + TR (Ours)} \\
\cmidrule(lr){5-7}\cmidrule(lr){8-10}\cmidrule(lr){11-13}\cmidrule(lr){14-16}
Model & Benchmark & $H$ & $1/H$ & Share & Cos & Gain & Share & Cos & Gain & Share & Cos & Gain & Share & Cos & Gain \\
\midrule
$\pi_{0.5}$ & LIBERO & 10 & 0.10 & 0.77 & 0.88 & 0.87 & 0.12 & 0.02 & 0.00 & 0.70 & 0.89 & 0.55 & 0.10 & 0.02 & 0.00 \\
GR00T N1.7 & LIBERO & 40 & 0.03 & 0.29 & 0.89 & 1.06 & 0.04 & $-0.01$ & 0.00 & 0.28 & 0.84 & 0.37 & 0.06 & $-0.01$ & 0.00 \\
GR00T N1.7 & Fractal & 40 & 0.03 & 0.05 & 0.96 & 1.19 & 0.00 & $-0.07$ & $-0.03$ & 0.03 & 0.39 & 0.38 & 0.02 & 0.03 & 0.02 \\
GR00T N1.7 & Bridge & 40 & 0.03 & 0.04 & 0.91 & 0.92 & 0.00 & $-0.01$ & 0.00 & 0.03 & 0.52 & 0.74 & 0.02 & 0.05 & 0.09 \\
\bottomrule
\end{tabular}
\end{table*}

\paragraph{Per-sample measurement.}
The decomposition predicts that, at a fixed condition, the directional bias of each sample follows the temporal mean of its noise.
To test this prediction, we report in \cref{tab:per_sample_models} the cosine between $b_c(\epsilon)$ and $\bar{\epsilon}$ for individual samples, without averaging action deviations over conditions.
For each variant, we measure directional bias relative to its own noise-averaged prediction.
We use 480 LIBERO conditions and 120 observations from each of Fractal and Bridge, with 64 noise samples and their negations.
All statistics are computed over the full generated chunk of $H$ steps.
``Cos'' is the cosine between $b_c(\epsilon)$ and the temporal mean $\bar{\epsilon}$ of the originally drawn initial noise, averaged over all (condition, noise) pairs.
``Gain'' is the least-squares slope of $b_c(\epsilon)$ on $\bar{\epsilon}$, the empirical counterpart of $g_c$.
``Share'' is the fraction of noise-induced deviation energy carried by the temporal mean of the deviation, which equals $1/H$ if no temporal direction is preferred.
For the baselines, the mean per-sample cosine lies between $0.88$ and $0.96$ across all four model--dataset pairs, and the cosine between the biases of $\epsilon$ and $-\epsilon$ lies between $-0.99$ and $-0.93$.
Both measurements are consistent with the per-sample decomposition and with noise negation.
The columns with TR serve as a consistency check: once the temporal mean of the initial noise is removed, the remaining bias should no longer follow it, and the cosine and the gain indeed fall to near zero.

The temporal mean of the deviation carries $77\%$ and $29\%$ of the deviation energy for $\pi_{0.5}$ and GR00T~N1.7 on LIBERO, far above the $1/H$ expected when all temporal directions contribute equally.
The share exceeds $1/H$ on every model--dataset pair, by a factor between $1.4$ (Bridge) and $11.6$ (GR00T~N1.7 on LIBERO).
For the SimplerEnv checkpoints of GR00T~N1.7, the directional bias follows $\bar{\epsilon}$ equally closely, but the share is $0.04$--$0.05$, so most of the deviation energy lies in other temporal modes.
Since TR removes only the share carried by the temporal mean, this is consistent with the modest reduction in $M$ from TR alone in \cref{tab:simpler} (from $5.59$ to $4.93$ on Fractal), and with AC accounting for most of the reduction on that benchmark.

On LIBERO, we also examine the effect of AC training.
For $\pi_{0.5}$, the cosine after AC training is $0.89$, so the direction is preserved, while the gain decreases from $0.87$ to $0.55$.
For GR00T~N1.7, the gain decreases from $1.06$ to $0.37$.
AC shrinks the aligned component, and TR removes what remains.

Under the ideal Gaussian model, the aligned term $g_c\bar{\epsilon}$ is present for every sample and every condition.
The measurements in \cref{tab:per_sample_models} show strong per-sample alignment between directional bias and the temporal mean of the initial noise for all evaluated baselines.
In the next section, we examine how AC and TR affect this term.

\section{Energy-Matched Controls}
\label{app:remove}

This section connects the two corrections to the decomposition of \cref{app:additional_theory}.
For TR, we confirm experimentally that the constant temporal direction, and not the removed energy, is what matters.
AC acts during training on the component of the prediction that changes sign with the noise, which contains the directional bias.

\label{app:tr_controls}
\begin{table}[t]
\centering
\small
\setlength{\tabcolsep}{5pt}
\caption{\textbf{Noise modifications on the $\pi_{0.5}$ baseline.} Success rates (\%) on LIBERO.}
\label{tab:tr_controls}
\resizebox{\columnwidth}{!}{%
\begin{tabular}{@{}lccc@{}}
\toprule
Initial noise & Energy & Success (\%) & $z$ \\
\midrule
Baseline ($\epsilon$) & 0\% & 93.8 & $-4.6$ \\
\midrule
Uniform scaling $0.95\epsilon$ & 10\% & 93.5 & $-4.9$ \\
Random direction $\perp\mathbf{1}$ & 10\% & 93.4 & $-5.0$ \\
DCT mode 1 & 10\% & 94.6 & $-3.4$ \\
DCT mode 2 & 10\% & 94.3 & $-3.9$ \\
DCT mode 4 & 10\% & 93.4 & $-5.1$ \\
DCT mode 9 & 10\% & 94.2 & $-3.9$ \\
TR ($\widetilde{\epsilon}$, constant direction) & 10\% & \underline{96.7} & -- \\
TR, energy restored & 0\% & \textbf{96.8} & $+0.2$ \\
\bottomrule
\end{tabular}}
\end{table}

\paragraph{Comparison at matched energy.}
\cref{tab:tr_controls} compares TR with modifications of the initial noise on the 2,000 LIBERO episodes.
``Energy'' denotes the removed share of the expected noise energy, and $z$ is the paired McNemar statistic against TR on identical episodes, where negative values indicate a lower success rate than TR.
The rows of $\epsilon$ and TR repeat \cref{tab:ablation_beta}.
All rows are evaluated on the same 2,000 episodes, and $z$ pairs each row with the TR row.
Each control in the second block of \cref{tab:tr_controls} removes approximately $1/H=10\%$ of the expected noise energy, either uniformly or along one non-constant temporal direction.
The largest change among the non-constant directions is obtained with DCT mode 1 ($+0.8$ points over the baseline), whereas removing the constant direction raises the success rate by 2.9 points.
Paired McNemar tests on identical episodes separate TR from every energy-matched control ($z\le-3.4$).
Rescaling the mean-centered noise by $\sqrt{H/(H-1)}$ restores its expected energy while keeping its temporal mean at zero, and retains the gain of TR (96.8\%).
Among the controls that remove approximately 10\% of the noise energy, TR gives the highest success rate.

\paragraph{Contribution of the constant direction.}
In the open-loop outputs of the baseline, the temporal mean of the noise accounts for 10\% of the noise energy but for 62\% of the noise-induced variance of the generated actions, whereas a random temporal direction of equal energy accounts for 3\%.
This is the empirical counterpart of the gain $g_c$ in \cref{app:additional_theory}.

\section{Beyond the Idealized Setting}
\label{app:robustness}

Our analysis assumes exact integration of the sampling ODE and a Gaussian action distribution.
In practice, the policies use a few Euler steps (10 for $\pi_{0.5}$ and 4 for GR00T~N1.7), and the action distribution can be multimodal.
We examine whether the relationship between the temporal mean of the initial noise and directional bias holds when each assumption is relaxed.
Both results below concern the optimal velocity field.
\cref{tab:per_sample_models} compares their predictions with learned policies.
\paragraph{Finite-step Euler sampling.}
We first replace exact integration with $N\ge2$ Euler steps of Eq.~\eqref{eq:fm_generation}, using the optimal velocity field of Eq.~\eqref{eq:app_optimal_velocity} at times $t_k=1-k/N$.
Under the Gaussian model, the generated action sequence remains linear in the initial noise:
\begin{equation}
y_0^{(N)}=\mu_c+S_N\epsilon,\quad
S_N=\prod_{k=0}^{N-1}\Bigl(I-\tfrac{1}{N}B_{t_k}C_{t_k}^{-1}\Bigr).
\label{eq:app_euler}
\end{equation}
To see this, we use the centered trajectory $W_t=y_t-(1-t)\mu_c$ of Eq.~\eqref{eq:app_centered_trajectory}.
Since the affine field tracks the mean term exactly, one Euler step from $t$ to $t-\delta$, with $\delta=1/N$, gives $W_{t-\delta}=(I-\delta B_tC_t^{-1})W_t$.
Iterating from $W_1=\epsilon$ yields Eq.~\eqref{eq:app_euler}.
Each factor is a function of $\Sigma_c$, so $S_N$ is symmetric and shares the eigenvectors of $\Sigma_c$.
For the mode with eigenvalue $\sigma_i$, the factor equals $[(1-t)\sigma_i(1-t+\delta)+t(t-\delta)]/c_i(t)$.
This value is $1-\delta>0$ at $t=1$, and it is positive for $\delta\le t<1$ because both terms in the numerator are nonnegative and the first is positive.
Hence, $S_N$ is positive definite.

The per-sample decomposition of Eq.~\eqref{eq:app_per_sample} therefore holds for the finite-step sampler with $\Sigma_c^{1/2}$ replaced by $S_N$.
The gain becomes $g_c^{(N)}=\tfrac{1}{H}\mathbf{1}^{\top}S_N\mathbf{1}$, which is again positive, and TR removes the aligned term exactly.
Finite-step sampling thus changes the gain and the residual, but not the form of the decomposition.

\paragraph{Non-Gaussian motion shape.}
We next relax the Gaussian assumption on the shape of the action sequence.
We split the target action sequence into its temporal mean and a mean-centered component:
\begin{equation}
x_0=\mathbf{1}\bar{a}^{\top}+A_\perp,
\qquad
\bar{a}=\tfrac{1}{H}x_0^{\top}\mathbf{1},
\qquad
A_\perp=Px_0.
\label{eq:app_action_split}
\end{equation}
We assume that, given $c$, the temporal mean is Gaussian, $\bar{a}\sim\mathcal{N}(\bar{m}_c,\tfrac{s_c^2}{H}I_d)$ with $s_c>0$, and independent of $A_\perp$.
The distribution of $A_\perp$ is otherwise arbitrary with finite second moments and may be multimodal.
Under these assumptions, the optimal velocity field generates, for every noise sample,
\begin{equation}
\begin{aligned}
\tfrac{1}{H}\bigl(y_0^\star\bigr)^{\top}\mathbf{1}&=\bar{m}_c+s_c\,\bar{\epsilon},\\
P\,y_0^\star(\widetilde{\epsilon})&=P\,y_0^\star(\epsilon).
\end{aligned}
\label{eq:app_nongaussian}
\end{equation}
The first identity states that the temporal mean of the generated actions follows $\bar{\epsilon}$ exactly, with gain $s_c$.
The second states that TR leaves the mean-centered component of the generated actions unchanged.

Both identities follow because the temporal mean and the mean-centered component evolve independently.
By the noise decomposition in \cref{app:additional_theory} and the assumed independence, $(\bar{a},\bar{\epsilon})$ is independent of $(A_\perp,\widetilde{\epsilon})$ given $c$, so the temporal mean and the mean-centered component of $x_t$ are conditionally independent.
The conditional expectation $\mathbb{E}[\epsilon-x_0\mid x_t,c]$ then splits into a term that depends only on the temporal mean of $x_t$ and a term that depends only on $Px_t$.
The temporal mean of the sampling trajectory thus obeys a closed linear ODE with a scalar gain, which is the ODE of the mode with eigenvalue $s_c^2$ in \cref{app:closed_form_generation}.
Its solution at $t=0$ is $\bar{m}_c+s_c\bar{\epsilon}$.
The mean-centered component evolves independently of $\bar{\epsilon}$, which gives the second identity.

The Gaussian model with $\mathbf{1}$ an eigenvector of $\Sigma_c$ is a special case of this setting.
Under this weaker assumption, the per-sample identity $b_c(\epsilon)=s_c\bar{\epsilon}$ holds, and TR leaves the mean-centered component unchanged even when its distribution is multimodal.
The motion shape represented by this component may thus follow any distribution.
The temporal mean of the action sequence must still be Gaussian and independent of the mean-centered component.
A multimodal temporal mean is not covered.
Beyond these theoretical extensions, we test AC and TR on a second backbone on a real robot in \cref{app:real_robot}.

\section{Additional Real-Robot Experiments}
\label{app:real_robot}

\subsection{Results on GR00T N1.7}
\label{app:real_groot}

We examine whether AC and TR improve real-robot performance across backbones.
The main paper evaluates $\pi_{0.5}$.
We repeat the full protocol with GR00T~N1.7, whose action expert and number of sampling steps differ from those of $\pi_{0.5}$.
We fine-tune GR00T~N1.7 on the same demonstrations (50 per task, 500 in total) for 60k steps with a batch size of 96. For AC, each backbone keeps the weight used in simulation ($\lambda=3$ for $\pi_{0.5}$ and $\lambda=5$ for GR00T~N1.7).
As for $\pi_{0.5}$, we evaluate four variants on the 10 tasks with five trials each.

\begin{table}[t]
\centering
\small
\setlength{\tabcolsep}{4pt}
\caption{\textbf{Real-robot success rates on two backbones.} Success rates (\%) over 50 trials with 95\% intervals.}
\label{tab:real_two_backbones}
\resizebox{\columnwidth}{!}{%
\begin{tabular}{@{}lcccc@{}}
\toprule
& \multicolumn{2}{c}{$\pi_{0.5}$} & \multicolumn{2}{c}{GR00T N1.7} \\
\cmidrule(lr){2-3}\cmidrule(lr){4-5}
Method & Success (\%) & $\Delta$ & Success (\%) & $\Delta$ \\
\midrule
Baseline & 40 {\scriptsize[28, 54]} & -- & 28 {\scriptsize[17, 42]} & -- \\
+ TR & 50 {\scriptsize[37, 63]} & $+10$ {\scriptsize[$-9$, $+28$]} & 38 {\scriptsize[26, 52]} & $+10$ {\scriptsize[$-8$, $+27$]} \\
+ AC & 48 {\scriptsize[35, 61]} & $+8$ {\scriptsize[$-11$, $+26$]} & 48 {\scriptsize[35, 61]} & $+20$ {\scriptsize[$+1$, $+37$]} \\
+ AC + TR (Ours) & \textbf{68} {\scriptsize[54, 79]} & $+28$ {\scriptsize[$+8$, $+45$]} & \textbf{60} {\scriptsize[46, 72]} & $+32$ {\scriptsize[$+13$, $+48$]} \\
\bottomrule
\end{tabular}}
\end{table}

\paragraph{Overall results.}
We report overall success rates over 50 trials (10 tasks $\times$ 5 trials) in \cref{tab:real_two_backbones}, with 95\% Wilson intervals.
$\Delta$ denotes the difference from the baseline of the same backbone in percentage points, with its 95\% Newcombe interval.
AC+TR increases the success rate of GR00T~N1.7 from 28\% (14/50) to 60\% (30/50).
This gain of 32 percentage points has a 95\% interval that excludes zero.
For $\pi_{0.5}$, success increases from 40\% to 68\%.
On both backbones, each correction alone exceeds the baseline, and AC+TR achieves the highest success rate.
The contributions differ across backbones.
For GR00T~N1.7, AC alone gives $+20$ percentage points over the baseline, with an interval that excludes zero, and AC+TR is 22 points above TR alone (60\% versus 38\%).
TR alone gives $+10$ percentage points over the baseline on both backbones.
For $\pi_{0.5}$, each correction alone gives $+8$ to $+10$ percentage points, while AC+TR gives $+28$.
The larger contribution of AC for GR00T~N1.7 is consistent with \cref{tab:simpler}, where AC accounts for most of the reduction in $M$.
With 50 trials per cell, the reported intervals include zero for TR alone on both backbones and for AC alone on $\pi_{0.5}$.
The intervals for AC+TR versus baseline exclude zero on both backbones.

\begin{figure*}[t]
\centering
\includegraphics[width=\textwidth]{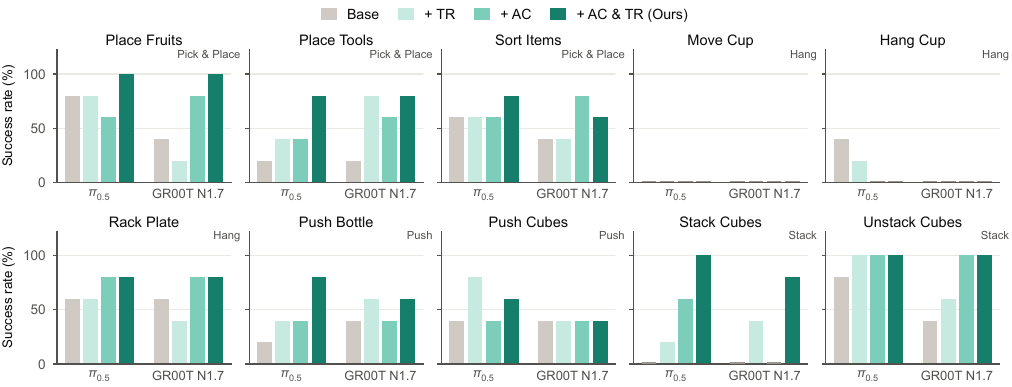}
\caption{\textbf{Per-task real-robot results.} Success rates (\%) for each task.}
\label{fig:real_per_task}
\end{figure*}

\paragraph{Per-task results.}
\cref{fig:real_per_task} breaks the success rates down by task for each variant and backbone, with task names and categories following \cref{fig:fig_7}.
AC+TR matches or exceeds the baseline on 9 of the 10 tasks for $\pi_{0.5}$ and on all 10 tasks for GR00T~N1.7.
The largest gains appear on Stack Cubes (0 to 5 successes for $\pi_{0.5}$ and 0 to 4 for GR00T~N1.7), Place Tools, and Push Bottle.
The two cup tasks of the Hang category (Move Cup and Hang Cup) remain essentially unsolved by every variant on both backbones, and the only task on which the baseline is ahead (Hang Cup, $\pi_{0.5}$, 2 versus 0) is one of them.
The corrections reduce the sensitivity to the initial noise but do not compensate for task behaviors that the policies have not learned.

\subsection{Execution Time}
\label{app:real_time}

\begin{table}[t]
\centering
\small
\setlength{\tabcolsep}{5pt}
\caption{\textbf{Execution time on pick-and-place tasks.} Mean duration (s) of successful trials.}
\label{tab:real_time}
\begin{tabular}{@{}lcc@{}}
\toprule
& \multicolumn{2}{c}{Time (s)} \\
\cmidrule(lr){2-3}
Method & $\pi_{0.5}$ & GR00T N1.7 \\
\midrule
Baseline & 34.5 & 43.3 \\
+ TR & 31.1 & 38.7 \\
+ AC & 32.1 & \textbf{32.5} \\
+ AC + TR (Ours) & \textbf{27.7} & 33.8 \\
\bottomrule
\end{tabular}
\end{table}

We next compare the execution time in a separate pick-and-place experiment in \cref{tab:real_time}.
For each method, we perform 10 success trials per task and measure the time to successful completion.
We compute durations from the number of 15\,Hz control steps until the operator stops the rollout upon success.
These durations include the operator's reaction time.
For $\pi_{0.5}$, AC+TR has a mean duration of 27.7\,s, compared with 34.5\,s for the baseline.
The mean is lower with TR both without AC and with AC (34.5 to 31.1\,s and 32.1 to 27.7\,s).
This is consistent with the baseline's difficulty recovering from inaccurate motions described in \cref{sec:real_exp}.
For GR00T N1.7, AC+TR is also faster than the baseline (33.8 versus 43.3\,s), while AC alone has the shortest mean duration of 32.5\,s.
These results show shorter mean execution times with AC+TR than with the baseline for both backbones on the evaluated pick-and-place tasks.

\subsection{Qualitative Results}
\label{app:real_qual}

\begin{figure*}[t]
\centering
\includegraphics[width=\textwidth]{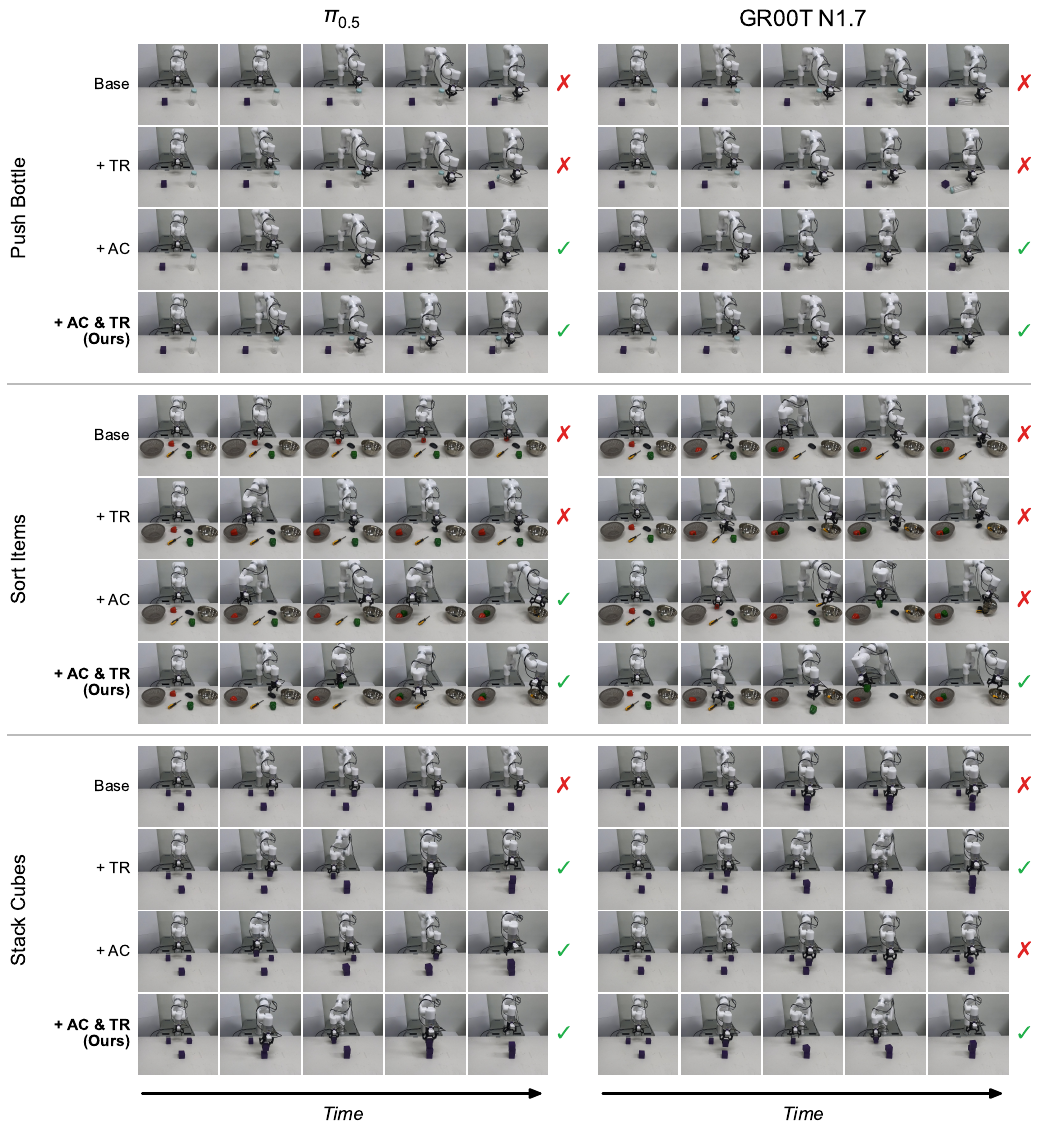}
\caption{\textbf{Real-robot rollouts on three tasks.} Check marks and crosses indicate success and failure.}
\label{fig:real_qual}
\end{figure*}

We further compare execution behaviors in \cref{fig:real_qual}, which shows one rollout per variant for $\pi_{0.5}$ (left) and GR00T~N1.7 (right) on Push Bottle, Sort Items, and Stack Cubes, with frames sampled uniformly in time.
In Push Bottle, the baseline and TR alone tip the bottle over on both backbones.
The two AC-trained variants keep it upright and push it next to the cube.
In Sort Items, the policy must place the fruits and the tools in separate containers.
For $\pi_{0.5}$, the two AC-trained variants complete the task.
For GR00T~N1.7, only AC+TR completes it.
In Stack Cubes, the baseline fails to place the cube on both backbones.
TR alone succeeds, and AC+TR succeeds in both rollouts shown.
Across the three tasks and the two backbones, AC+TR is the only variant that succeeds in all six rollouts shown.
This is consistent with the success rates in \cref{tab:real_two_backbones}.
These examples illustrate execution behavior and do not replace the aggregate counts in \cref{tab:real_two_backbones}.

\section{Implementation Details}
\label{app:impl}

\paragraph{Action chunks and sampling.}
\cref{tab:impl_settings} lists the number of predicted actions per chunk ($H$), the number of Euler denoising steps (Steps), and the number of actions executed before replanning (Exec.) for every setting.
All policies use the Euler sampler of their official code base.
The policy replans after executing the listed number of actions, and a new noise sample is drawn at every replanning step.
On LIBERO the end-effector commands are per-step relative actions, and on the real robot they are relative to the state at the start of the chunk.

\begin{table}[t]
\centering
\small
\setlength{\tabcolsep}{4pt}
\caption{\textbf{Chunking and sampling settings.} Settings for each model and benchmark.}
\label{tab:impl_settings}
\begin{tabular}{@{}llccc@{}}
\toprule
Model & Benchmark & $H$ & Steps & Exec. \\
\midrule
$\pi_{0.5}$ & LIBERO, LIBERO-Plus & 10 & 10 & 5 \\
GR00T N1.7 & LIBERO, LIBERO-Plus & 40 & 4 & 8 \\
GR00T N1.7 & SimplerEnv WidowX & 40 & 4 & 4 \\
GR00T N1.7 & SimplerEnv Google Robot & 40 & 4 & 1 \\
\bottomrule
\end{tabular}
\end{table}

\paragraph{Temporal mean removal.}
TR is applied to the initial noise of every inference call, to all action dimensions including the gripper and padded dimensions, by subtracting the mean over the $H$ steps of the chunk (Eq.~\eqref{eq:temporal_mean_removal}).
The sampler, the number of steps, and the checkpoint are unchanged, and no additional network evaluation is required.
The baseline and the baseline with TR use the same checkpoint, as do AC and AC+TR.
TR is not used during training.

\paragraph{Antithetic noise consistency.}
The two branches of Eq.~\eqref{eq:inverse_noise_consistency} share the observation, the instruction, the target $x_0$, and the timestep $t$, which is sampled from the distribution of the original training code.
The vision-language prefix is computed once and only the action expert is evaluated twice.
The flow-matching loss is computed on the $\epsilon$ branch and the gradient of the AC term flows through both branches.
For $\pi_{0.5}$, a training step with AC takes 1.7 times as long as a baseline step on the same GPU (4.5 versus 2.6 seconds per step).
The cost is confined to training.

\paragraph{Evaluation.}
All variants of a model are evaluated on identical episodes.
LIBERO uses the 50 standard initial states of each of the 40 tasks, and LIBERO-Plus all 10,030 perturbation instances without additional training.
$M$ and $C$ (Eqs.~\eqref{eq:noise_induced_variance} and \eqref{eq:cross_condition_consistency}) are computed open loop in the normalized action space of the model, with 64 noise samples shared across variants for each of 120 observations per dataset.
$\overline{y}_0(c)$ is the average over these samples.
\cref{tab:libero_results} combines results measured by us (marked $\dagger$) with numbers reported by the cited papers for other backbones.
Only the rows that share a backbone are controlled comparisons.

\paragraph{Paired tests.}
All variants of a model are evaluated on the same episodes (the same task, initial state, and perturbation instance), so differences can be tested with McNemar's test on paired outcomes.
For the LIBERO-Plus results of $\pi_{0.5}$ in \cref{tab:ablation_beta}, TR improves the baseline from 71.8\% to 75.3\% ($+3.5\pm0.8$ points, $z=8.8$, with 973 episodes solved only with TR and 623 only without) and the AC model from 75.6\% to 77.3\% ($+1.7\pm0.7$ points, $z=4.9$, 700 versus 527) on the same 10,030 episodes (95\% intervals).

\end{document}